\documentclass{article}

\PassOptionsToPackage{numbers,compress,sort}{natbib}

\usepackage[preprint]{neurips_2021}

\usepackage[utf8]{inputenc} 
\usepackage[T1]{fontenc}    
\usepackage{hyperref}       
\usepackage{url}            
\usepackage{booktabs}       
\usepackage{amsfonts}       
\usepackage{nicefrac}       
\usepackage{microtype}      
\usepackage{xcolor}         

\usepackage{amssymb}
\usepackage{amsmath}
\newtheorem{theorem}{Theorem}
\newtheorem{corollary}{Corollary}
\usepackage{graphicx}
\usepackage{subcaption}
\usepackage{multirow}
\usepackage{wrapfig}

\title{Learning Conjoint Attentions for Graph Neural Nets}

\author{%
  Tiantian~He$^{\textbf{1,2}}$~~~Yew-Soon~Ong$^{\textbf{1,2}}$~~~Lu Bai$^{\textbf{1,2}}$\\
  $^1$Agency for Science, Technology and Research (A*STAR)\\
  $^2$DSAIR, Nanyang Technological University\\
  \texttt{$\lbrace$He\_Tiantian,Bai\_Lu$\rbrace$@ihpc.a-star.edu.sg, Ong\_Yew\_Soon@hq.a-star.edu.sg}\\
  \texttt{$\lbrace$tiantian.he,bailu,asysong$\rbrace$@ntu.edu.sg}\\
}

\begin{document}

\maketitle

\begin{abstract}
	In this paper, we present Conjoint Attentions (CAs), a class of novel learning-to-attend strategies for graph neural networks (GNNs).
	Besides considering the layer-wise node features propagated within the GNN, CAs can additionally incorporate various structural interventions, such as node cluster embedding, and higher-order structural correlations that can be learned outside of GNN, when computing attention scores.
	The node features that are regarded as significant by the conjoint criteria are therefore more likely to be propagated in the GNN.
	Given the novel Conjoint Attention strategies, we then propose Graph conjoint attention networks (CATs) that can learn representations embedded with significant latent features deemed by the Conjoint Attentions.
	Besides, we theoretically validate the discriminative capacity of CATs. 
	CATs utilizing the proposed conjoint attention strategies have been extensively tested in well-established benchmarking datasets and comprehensively compared with state-of-the-art baselines.
	The obtained notable performance demonstrates the effectiveness of the proposed conjoint attentions.
\end{abstract}

\section{Introduction}\label{intro}
Graph neural networks (GNNs) have shown much success in the learning of graph structured data.
Amongst these noteworthy GNNs, attention-based GNNs \cite{velivckovic2018graph} have drawn increasing interest lately, and have been applied to solve a plethora of real-world problems competently, including node classification \cite{velivckovic2018graph,kipf2016semi}, image segmentation \cite{wang2019graph}, and social recommendations \cite{song2019session}.

Empirical attention mechanisms adopted by GNNs aim to leverage the node features (node embeddings) to compute the normalized correlations between pairs of nodes that are observed to connect.
Treating normalized correlations (attention scores/coefficients) as the relative weights between node pairs, attention-based GNN typically performs a weighted sum of node features which are subsequently propagated to higher layers.
Compared with other GNNs, especially those that aggregate node features with predefined strategies \cite{kipf2016semi,atwood2016diffusion,klicpera2019diffusion}, attention-based GNNs provide a dynamical way for feature aggregation, which enables highly correlated features from neighboring nodes to be propagated in the multi-layer neural architecture.
Representations that embed with multi-layer correlated features are consequently learned by attention-based GNNs, and can be used for various downstream tasks.

Though effective, present empirical graph attention has several shortcomings when aggregating node features.
First, the computation of attention coefficients is limited solely to the correlations of internal factors, i.e., layer-wise node features within the neural nets.
External factors such as cluster structure and higher-order structural similarities, which comprise heterogeneous node-node relevance have remained underexplored to be positively incorporated into the computation of more purposeful attention scores.
Second, the empirical attention heavily leaning on the node features may cause over-fitting in the training stage of neural nets \cite{wang2019improving}.
The predictive power of attention-based GNNs is consequently limited.


To overcome the mentioned challenges, in this paper, we propose a class of generic graph attention mechanisms, dubbed here as Conjoint Attentions (CAs).
Given CAs, we construct Graph conjoint attention networks (CATs) for different downstream analytical tasks.
Different from previous graph attentions, CAs are able to flexibly compute the attention coefficients by not solely relying on layer-wise node embeddings, but also allowing the incorporation of purposeful interventions brought by factors external to the neural net, e.g., node cluster embeddings.
With this, CATs are able to learn representations from features that are found as significant by diverse criteria, thus increasing the corresponding predictive power.
The main contributions of the paper are summarized as follows.
\begin{itemize}
	\item We propose Conjoint Attentions (CAs) for GNNs.
	Different from popular graph attentions that rely solely on node features, CAs are able to incorporate heterogeneous learnable factors that can be internal and/or external to the neural net to compute purposeful and more appropriate attention coefficients.
	The learning capability and hence performance of CA-based GNNs is thereby enhanced with the proposed novel attention mechanisms.
	\item For the first time, we theoretically analyze the expressive power of graph attention layers considering heterogeneous factors for node feature aggregation, and the discriminant capacity of such attention layers, i.e., CA layers is validated.
	\item Given CA layers, we build and demonstrate the potential of Graph conjoint attention networks (CATs) for various learning tasks.
	The proposed CATs are comprehensively investigated on established and extensive benchmarking datasets with comparison studies to a number of state-of-the-art baselines.
	The notable results obtained are presented to verify and validate the effectiveness of the newly proposed attention mechanisms.
\end{itemize}

\section{Related works}\label{related-works}
To effectively learn low-dimensional representations in graph structured data, many GNNs have been proposed to date.
According to the ways through which GNNs define the layer-wise operators for feature aggregation, GNNs can generally be categorized as spectral or spatial \cite{wu2020comprehensive}.

\textbf {Spectral GNNs}-The layer-wise function for feature aggregation in spectral GNNs is defined according to the spectral representation of the graph.
For example, Spectral CNN \cite{bruna2014spectral} constructs the convolution layer based on the eigen-decomposition of graph Laplacian in the Fourier domain.
However, such layer is computationally demanding.
To reduce such computational burden, several approaches adopting the convolution operators which are based on simplified or approximate spectral graph theory have been proposed.
First, parameterized filters with smooth coefficients are introduced for Spectral CNN to incorporate spatially localized nodes in the graph \cite{henaff2015deep}.
Chebyshev expansion \cite{defferrard2016convolutional}  is then introduced to approximate graph Laplacian rather than directly performing eigen-decomposition.
Finally, the graph convolution filter is further simplified by only considering first or higher order of connected neighbors \cite{kipf2016semi,wu2019simplifying}, so as to make the convolution layer more computationally efficient.

\textbf {Spatial GNNs}-In contrast, spatial GNNs define the convolution operators for feature aggregation by directly making use of local structural properties of the central node.
The essence of spatial GNNs consequently lies in designing an appropriate function for aggregating the effect brought by the features of candidate neighbors selected based on appropriate sampling strategy. 
To achieve this, it sometimes requires to learn a weight matrix that accords to the node degree \cite{duvenaud2015convolutional}, utilize the power of transition matrix to preserve neighbor importance \cite{atwood2016diffusion,busch2020pushnet,klicpera2019diffusion,xu2018representation,klicperapredict}, extract the normalized neighbors \cite{niepert2016learning}, or sample a fixed number of neighbors \cite{hamilton2017inductive, zhang2019adaptive}.

As representative spatial GNNs, attention-based GNNs (GATs) \cite{velivckovic2018graph,gulcehre2018hyperbolic} have shown promising performances on various learning tasks.
What makes them effective in graph learning is a result of adopting the attention mechanism, which has been successfully used in machine reading and translation \cite{cheng2016long,luong2015effective}, and video processing \cite{xu2015show}, to compute the node-feature-based attention scores between a central node and its one-hop neighbors (including the central node itself).
Then, attention-based GNNs use the attention scores to obtain a weighted aggregation of node features which are subsequently propagated to the next layer.
As a result, those neighbors possessing similar features may then induce greater impact on the center node, and meaningful representations can be inferred by GATs.
Having investigated previous efforts to graph neural networks, we observe that the computation of empirical graph attentions heavily relies on layer-wise node features, while other factors, e.g., structural properties that can be learned outside of neural net, have otherwise been overlooked.
This motivates us in proposing novel attention mechanisms in this paper to alleviate the shortcomings of existing attention-based graph neural networks. 

\section{Graph conjoint attention networks}
In this section, we elaborate the proposed Conjoint Attention mechanisms, which are the cornerstones for building layers of novel attention-based graph neural networks.
Mathematical preliminaries and notations used in the paper are firstly illustrated.
Then, how to construct neural layers utilizing various Conjoint Attention mechanisms are introduced.
Given the formulated Conjoint Attention layers, we finally construct the Graph conjoint attention networks (CATs).

\subsection{Notations and preliminaries}
Throughout this paper, we assume a graph $G = \lbrace V, E \rbrace$ containing $N$ nodes, $|E|$ edges, and $C$ classes ($C\ll N$) to which the nodes belong, where $V$ and $E$ respectively represent the node and edge set.
We use $\mathbf A \in \lbrace 0, 1\rbrace^{N \times N}$ and $\mathbf X \in \mathbb R^{N \times D}$ to represent graph adjacency matrix and input node feature matrix, respectively.
$\mathcal N_i$ denotes the union of node $i$ and its one-hop neighbors.
$\mathbf W^l$ and $\lbrace \mathbf h^l_i \rbrace_{i = 1, ... N}$ denote the weight matrix and features (embeddings) of node $i$ at $l$th layer of CATs, respectively, and $\mathbf h^0$ is set to be the input feature, i.e., $\mathbf X$.
For the nodes in $\mathcal N_i$, their possible feature vectors form a multiset $M_i = (S_i, \mu_i)$, where $S_i = \lbrace s_1, ... s_n\rbrace$ is the ground set of $M_i$ which contains the distinct elements existing in $M_i$, and $\mu_i : S_i \rightarrow \mathbb N^\star$ is the multiplicity function indicating the frequency of occurrence of each distinct $s$ in $M_i$.  

\subsection{Structural interventions for Conjoint Attentions}\label{enlight}
As aforementioned, the proposed Conjoint Attentions are able to make use of factors that are either internal or external to the neural net to compute new attention coefficients.
Internal factors refer the layer-wise node embeddings in the GNN.
While, the external factors include various parameters that can be learned outside of the graph neural net and can potentially be used to compute the attention scores.
Taking the cue from cognitive science, where contextual interventions have been identified as effective external factors that may improve the attention and cognitive abilities \cite{jones2004joint}, here we refer these external structural properties as \textit {structural interventions} henceforth for the computing of attention coefficients.
Next, we propose a simple but effective way for CAs to capture diverse structural interventions external to the GNNs.
Let $\mathbf {C}_{ij}$ be some structural intervention between $i$th and $j$th node in the graph. It can be obtained with the following generic generating function:
\begin{equation}\label{gen}
	\begin{aligned}
		\mathbf {C}_{ij} = \mathop{\arg \min}_{\phi(\mathbf {C})_{ij}} \Psi(\phi(\mathbf {C})_{ij},\mathbf Y_{ij}),
	\end{aligned}
\end{equation}
where $\Psi(\cdot)$ represents a distance function and $\phi(\cdot)$ stands for an operator transforming $\mathbf {C}$ to the same dimensionality of $\mathbf Y$.
Given the generic generating function in Eq. (\ref{gen}), it is known that many effective paradigms for learning latent features can be used for the subsequent computation of conjoint attentions, if the prior feature matrix $\mathbf Y$ is appropriately provided.
Taking $\mathbf {A}$ as the prior feature matrix, in this paper, we consider two generation processes that can capture two unique forms of structural interventions.
Let $\Psi(\cdot)$ be the euclidean distance, when $\phi(\mathbf {C})_{ij} \doteq \mathbf V \mathbf V^{T}_{ij}$, we have:
\begin{equation}\label{local}
	\mathbf {C}_{ij} = \mathop{\arg \min}_{\mathbf V \mathbf V^{T}_{ij}}(\mathbf A_{ij} - \mathbf V \mathbf V^{T}_{ij})^2,
\end{equation}
where we use an $N$-by-$C$ matrix $\mathbf V$ to approximate $\mathbf {C}$ to reduce the computational burden.
As it shows in Eq. (\ref{local}), $\mathbf {C}_{ij}$ attempts to acquire the structural correlation pertaining to node cluster embeddings, based on matrix factorization (MF).
A higher $\mathbf {C}_{ij}$ learned by Eq. (\ref{local}) means a pair of nodes are very likely to belong to the same cluster.
If $\phi(\mathbf {C})_{ij} \doteq \sum_{j}\mathbf V \mathbf V^{T}_{ij}\mathbf A_{ij}$, we have:
\begin{equation}\label{global}
	\mathbf {C}_{ij} = \mathop{\arg \min}_{\mathbf V \mathbf V^{T}_{ij}}(\mathbf A_{ij} - \sum_{j}\mathbf V \mathbf V^{T}_{ij}\mathbf A_{ij})^2.
\end{equation}
As shown in Eq. (\ref{global}), $\mathbf {C}_{ij}$ is the coefficient of self-expressiveness \cite{elhamifar2013sparse} (SC) which may describe the global relation between node $i$ and $j$.
A higher $\mathbf {C}_{ij}$ inferred by Eq. (\ref{global}) means the global structure of node $i$ can be better represented by that of node $j$, and consequently this pair of nodes are more structurally correlated.
It is known that both aforementioned properties have not been considered previously by empirical graph attention mechanisms.
We believe considering either of them as structural interventions for the Conjoint Attentions could lead to better attention scores for feature aggregation.
Note that other types of $\mathbf {C}$ may also be feasible for the proposed attention mechanisms, as long as they are able to capture meaningful property which is not already possessed within the node embeddings of the GNN.

\begin{figure}[t]
	\centering
	\includegraphics[width=0.7\textwidth]{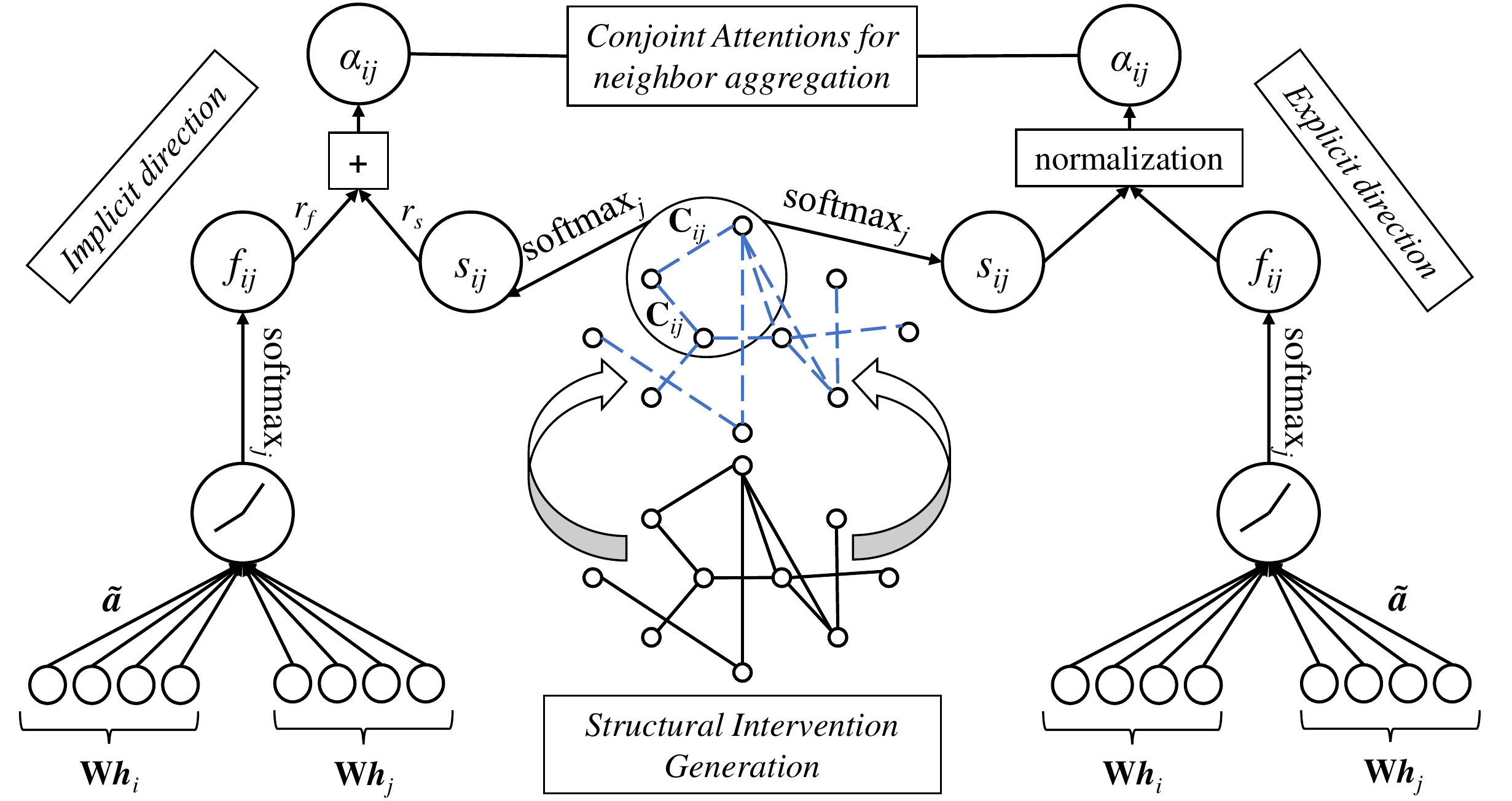}
	\caption{Graphical illustration of the Conjoint Attention layer used in CATs. Left: CA mechanism using \textit{Implicit direction} strategy (CAT-I). Right: CA mechanism using \textit{Explicit direction} strategy (CAT-E). Both two mechanisms consider learnable structural interventions.}\label{attm}
\end{figure}

\subsection{Conjoint attention layer}
Having obtained a proper $\mathbf {C}$, we present next the Conjoint Attention layer, which is the core module for building CATs and will be used in our experimental study. 
Different from the attention layers considered in other GNNs, the attention layer proposed here adopts novel attention mechanisms, i.e., Conjoint Attentions.
It is known that empirical graph attentions solely concern the correlations pertaining to the internal factors, e.g., node embeddings locating in each layer of the neural net.
How those diverse forms of relevance, e.g., the correlation in terms of node cluster embeddings and self-expressiveness may affect the representation learning is therefore yet to be investigated in previous works.
Besides utilizing the correlations pertaining to node embeddings, the proposed Conjoint Attentions now take into additional considerations the structural interventions brought by diverse node-node relevance, which is learned outside of the neural network. 
As a result, each Conjoint Attention layer may pay more attentions to the similar embeddings of neighbors, as well as to the ones that share other forms of relevance with the central node.
Node representations possessing heterogeneous forms of relevance can now be learned by the CATs.

Given a set of node features $\lbrace \mathbf h^l_i\rbrace_{i = 1, ... N}$, each $\mathbf h^l_i \in R^{D^l}$, the Conjoint Attention layer maps them into $D^{l+1}$ dimensional space $\lbrace \mathbf h^{l+1}_i\rbrace_{i = 1, ... N}$, according to the correlations of node features and aforementioned structural interventions.
The contextual correlation between two connected nodes, say $v_i$ and $v_j$ is firstly obtained.
To do so, we directly adopt the feature-based attention mechanism considered by existing graph attention networks (GAT) \cite{velivckovic2018graph}:
\begin{equation}\label{f-att}
	f_{ij}=\frac{\exp(\text{LeakyReLU}(\mathbf {\vec{a}}^T(\mathbf {W}^l\mathbf h^l_i\parallel \mathbf {W}^l\mathbf h^l_j )))}{\sum_{k\in \mathcal{N}_i} \exp(\text{LeakyReLU}(\mathbf {\vec{a}}^T(\mathbf {W}^l\mathbf h^l_i\parallel \mathbf {W}^l\mathbf h^l_k )))},
\end{equation}
where $\mathbf {\vec{a}}\in \mathbb{R}^{2D^{l+1}}$ is a vector of parameters of the feedforward layer, $\parallel$ stands for the concatenation function, and $\mathbf W^l$ is $D^{l+1}\times D^l$ parameter matrix for feature mapping.
Given Eq. (\ref{f-att}), the proposed CA layer captures the feature correlations between connected nodes (first-order neighbors) by computing the similarities w.r.t. node features mapped to next layer.

As mentioned, determining the attention scores solely based on node features internal to a GNN may result in overlooking other important factors.
To overcome this issue, the proposed CA layer attempts to learn the structural interventions as described in Section \ref{enlight}, and presents the additional new information for computing the attention scores.
Given any learnable parameter $\mathbf C_{ij}$ (structural intervention) between two nodes, CA layer additionally obtains a supplementary correlation as follows:
\begin{equation}
	s_{ij} = \frac{\exp{(\mathbf C_{ij})}}{\sum_{k \in \mathcal N_i} \exp{(\mathbf C_{ik})}}.
\end{equation}
Given $f_{ij}$ and $s_{ij}$, we propose two different strategies to compute the Conjoint Attention scores, aiming at allowing CATs to depend on the structural intervention at different levels.
The first mechanism is referred here as \textit {Implicit direction}.
It aims at computing the attention scores whose relative significance between structural and feature correlations can be automatically acquired.
To do so, each CA layer introduces two learnable parameters, $g_f$ and $g_s$, to determine the relative significance between feature and structural correlations and they can be obtained as follows:
\begin{equation}\label{pen}
	r_f = \frac{\exp{(g_f)}}{\exp{(g_s)+\exp{(g_f)}}}, r_s = \frac{\exp{(g_s)}}{\exp{(g_s)+\exp{(g_f)}}},
\end{equation}
where $r_s$ or $r_f$ represents the normalized significance related to different types of correlations. Given them, CAT then computes the attention score based on \textit {Implicit direction} strategy:
\begin{equation}\label{att-ip}
	\alpha_{ij} = \frac{r_f\cdot f_{ij}+r_s\cdot s_{ij}}{\sum_{k \in \mathcal N_i}[r_f\cdot f_{ik}+r_s\cdot s_{ik}]}=r_f\cdot f_{ij}+r_s\cdot s_{ij}.
\end{equation}
Given the attention mechanism shown in Eq. (\ref{att-ip}), $\alpha_{ij}$ attempts to capture the weighted mean attention in terms of the various node-node correlations, which are the ones internal or external to the GNN.
Compared with the attention mechanism solely based on features of one-hop neighbors, $\alpha_{ij}$ computed by Eq. (\ref{att-ip}) may be adapted according to the implicit impact brought by different structural interventions, e.g., correlations pertaining to node cluster embeddings and self-expressiveness coefficients.
Moreover, the relative significance $r$ can also be automatically inferred through the back propagation process.
More smooth and appropriate attention scores can thereby be computed by the CA layer for learning meaningful representations.

To enhance the impact of structural intervention, the CA layer has another strategy, named here as \textit{Explicit direction}, to compute attention scores between neighbors.
Given $f_{ij}$ and $s_{ij}$, the attention scores obtained via the \textit{Explicit direction} strategy is defined as follows:
\begin{equation}\label{att-dp}
	\alpha_{ij} = \frac{f_{ij}\cdot s_{ij} }{\sum_{k \in \mathcal N_i}f_{ik}\cdot s_{ik}}.
\end{equation}
Compared with Eq. (\ref{att-ip}), $s_{ij}$ explicitly influences the magnitude of $f_{ij}$, so that those node pairs which are irrelevant in terms of $\mathbf C_{ij}$ will never be assigned with high attention weights.
Based on \textit{Explicit direction} strategy, the CA layer becomes more structurally dependent when performing message passing to the higher layers in the neural architecture.

Having obtained the Conjoint Attention scores, the CA layer is now able to compute a linear combination of features corresponding to each node and its neighbors as output, which will be either propagated to the higher layer, or be used as the final representations for subsequent learning tasks.
The described output features can be computed as follows:
\begin{equation}\label{att-aggregation}
	\mathbf h^{l+1}_i = (\alpha_{ii}+\epsilon\cdot\frac{1}{\vert \mathcal N_i \vert})\mathbf {W}^l\mathbf h^l_i+ \sum_{j \in \mathcal N_i, j \ne i} \alpha_{ij} \mathbf {W}^l\mathbf h^l_j,
\end{equation}
where $\epsilon \in (0, 1)$ is a learnable parameter that improves the expressive capability of the proposed CA layer.

\subsection{Construction of Graph conjoint attention networks (CATs)}
In Fig. \ref{attm}, the Conjoint Attention layers that use the attention mechanisms proposed in this paper are graphically illustrated.
We are now able to construct Graph conjoint attention networks (CATs) using a particular number of CA layers proposed.
In practice, we also adopt the multi-head attention strategy \cite{vaswani2017attention} to stabilize the learning process.
CATs may either concatenate the node features from multiple hidden layers as the input for next layer, or compute the average of node features obtained by multiple units of output layers as the final node representations.
For the details on implementing multi-head attention in graph neural networks, the reader is referred to \cite{velivckovic2018graph}.

\section{Theoretical analysis}\label{theory}
Study on the expressive power of various GNNs has drawn much attention in the recent.
It concerns whether a given GNN can distinguish different structures where vertices possessing various vectorized features. 
It has been found that what the neighborhood aggregation functions of all message-passing GNNs aim at are analogous to the 1-dimensional Weisfeiler-Lehman test (1-WL test), which is injective and iteratively operated in the Weisfeiler-Lehman algorithm \cite{weisfeiler1968reduction, xu2018powerful, zhang2020improving}, does.
As a result, all message-passing GNNs are as most powerful as the 1-WL test \cite{xu2018powerful}.
The theoretical validation of the expressive power of a given GNN thereby lies in whether those adopted aggreation/readout functions are homogeneous to the 1-WL test.

One may naturally be interested in whether the expressive power of the proposed CAT layers is as powerful as the 1-WL test, which can distinguish all different graph structures.
To do so, we firstly show that neighborhood aggregation function (Eq. (\ref{att-aggregation})) without the term for improving expressive capability (i.e., $\epsilon\cdot\frac{1}{\vert \mathcal N_i \vert} \mathbf {Wh}^l_i$ in Eq. (\ref{att-aggregation})) still fails to discriminate some graph structures possessing certain topological properties.
Then, by integrating the term of improving expressive capability, all the proposed CA layers are able to distinguish all those graph structures that cannot be discriminated previously.

For the function of neighborhood aggregation solely utilizing the strategy shown in Eq. (\ref{att-ip}), we have the following theorem pointing out the conditions under which the aggregation function fails to distinguish different structures.
\begin{theorem}\label{theorem-ip}
	Assume the feature space $\mathcal X$ is countable and the aggregation function using the weights computed by Eq. (\ref{att-ip}) is represented as $h(c, X) = \sum_{x\in X} \alpha_{cx} g(x)$, where $c$ is the feature of center node, $X \in \mathcal X$ is a multiset containing the feature vectors from nodes in $\mathcal N_i$, $g(\cdot)$ is a function for mapping input feature $X$, and $\alpha_{cx}$ is the weight between $g(c)$ and $g(x)$. For all $g$ and the strategy in Eq. (\ref{att-ip}), $h(c_1, X_1) = h(c_2, X_2)$ if and only if $c_1 = c_2$, $X_1 = \lbrace S, \mu_1 \rbrace$, $X_2 = \lbrace S, \mu_2 \rbrace$, and $\sum_{y=x, y\in X_1} f_{c_1y}-\sum_{y=x, y\in X_2} f_{c_2y} = q[\sum_{y=x, y\in X_2} s_{c_2y} - \sum_{y=x, y\in X_1} s_{c_1y}]$, for $q = \frac{r_s}{r_f}$ and $x \in S$. In other words, $h$ will map different multisets into the same embedding iff the multisets have same central node feature, same underlying set, and the difference in feature-based scores is proportional ($ \frac{r_s}{r_f}$) to the opposite of that in the weights corresponding to the structural interventions.
\end{theorem}
We leave the proof of all the theorems and corollaries in the appendix.
For the aggregation function utilizing the strategy shown in Eq. (\ref{att-dp}), we have the following theorem indicating the structures which cannot be correctly distinguished. 
\begin{theorem}\label{theorem-dp}
	Under the same assumptions shown in Theorem \ref{theorem-ip}, for all $g$ and the strategy in Eq. (\ref{att-dp}), $h(c_1, X_1) = h(c_2, X_2)$ if and only if $c_1 = c_2$, $X_1 = \lbrace S, \mu_1 \rbrace$, $X_2 = \lbrace S, \mu_2 \rbrace$, and $q\cdot\sum_{y=x, y\in X_1} \phi(\mathbf C_{c_1x}) = \sum_{y=x, y\in X_2} \phi(\mathbf C_{c_2y})$, for $q > 0$ and $x \in S$, where $\phi(\cdot)$ is an function for mapping values to $\mathbb R^+$.
	In other words, $h$ will map different multisets into the same embedding iff the multisets have same central node feature, same node features whose corresponding mapped structural interventions are proportional.
\end{theorem}
Theorems \ref{theorem-ip} and \ref{theorem-dp} indicate that the CA layers may still fail to distinguish some structures, if they exclude the improving term shown in Eq. (\ref{att-aggregation}).
However, GNNs utilizing Eqs. (\ref{att-ip}) or (\ref{att-dp}) can still be more expressively powerful than classical GATs. 
As node features and structural interventions are heterogeneous, intuitively, structures satisfying the stated conditions should be infrequent.
This may well explain why those GNNs concerning including external factors, e.g., some structural properties into the computation of attention coefficients may experimentally perform better than GATs.
However, when distinct multisets with corresponding properties meet the conditions mentioned in Theorems \ref{theorem-ip} and \ref{theorem-dp}, the attention mechanisms solely based on Eqs. (\ref{att-ip}) or (\ref{att-dp}) cannot correctly distinguish such multisets.
Thus, GNNs only utilizing Eqs. (\ref{att-ip}) or (\ref{att-dp}) as the feature aggregation function fail to reach the upper bound of expressive power of all message-passing GNNs, i.e., the 1-WL test.

However, we are able to readily improve the expressive power of CATs to meet the condition of the 1-WL test by slightly modifying the aggregation function as Eq. (\ref{att-aggregation}) shows.
Then, the newly obtained Conjoint Attention scores can be used to aggregate the node features passed to the higher layers.
Next, we prove that the proposed Conjoint Attention mechanisms (Eqs. (\ref{att-ip})-(\ref{att-aggregation})) reach the upper bound of message-passing GNNs via showing they can distinguish those structures possessing the properties mentioned in Theorems \ref{theorem-ip} and \ref{theorem-dp}.
\begin{corollary}\label{coro-att}
	Let $\mathcal T$ be the attention-based aggregator shown in Eq. (\ref{att-aggregation}) that considers one of the strategies in Eq. (\ref{att-ip}) or (\ref{att-dp}) and operates on a multiset $H \in \mathcal H$, where $\mathcal H$ is a node feature space mapped from the countable input feature space $\mathcal X$.
	A $\mathcal H$ exists so that utilizing attention-based aggregator shown in Eq. (\ref{att-aggregation}), $\mathcal T$ can distinguish all different multisets in aggregation that it previously cannot discriminate.
\end{corollary}

Based on the performed analysis, the expressive power of CATs is theoretically stronger than state-of-the-art attention-based GNNs, e.g., GATs \cite{velivckovic2018graph}. 


\section{Experiments and analysis}\label{exp}
In this section, we evaluate the proposed Graph conjoint attention networks against a variety of state-of-the-art and popular baselines, on widely used network datasets.

\subsection{Experimental set-up}
\textbf{Baselines for comparison}-To validate the effectiveness of the proposed CATs, we compare them with a number of state-of-the-art baselines, including Arma filter GNN (ARMA) \citep{bianchi2021graph}, Simplified graph convolutional Networks (SGC) \citep{wu2019simplifying}, Personalized Pagerank GNN (APPNP) \citep{klicperapredict}, Graph attention networks (GAT) \citep{velivckovic2018graph}, Jumping knowledge networks (JKNet) \citep{xu2018representation}, Graph convolutional networks (GCN) \citep{kipf2016semi}, GraphSAGE \citep{hamilton2017inductive}, Mixture model CNN (MoNet) \citep{monti2017geometric}, and Graph isomorphism network (GIN) \cite{xu2018powerful}.
As GAT can alternatively consider graph structure by augmenting original node features (i.e., $\mathbf X$) with structural properties, we use prevalent methods for network embedding, including $k$-eigenvectors of graph Laplacian ($k$-Lap) \cite{qiu2018network}, Deepwalk \cite{perozzi2014deepwalk}, and Matrix factorization-based network embedding (NetMF) \cite{qiu2018network} to learn structural node representations and concatenate them with $\mathbf X$ as the input feature of GAT.
Thus, three variants of GAT, i.e., GAT-$k$-Lap, GAT-Deep, and GAT-NetMF are additionally constructed as compared baselines.
Based on the experimental results previously reported, these baselines may represent the most advanced techniques for learning in graph structured data.

\textbf{Testing datasets}-Five widely-used network datasets, which are Cora, Cite, Pubmed \cite{lu2003link,sen2008collective}, CoauthorCS \cite{shchur2018pitfalls}, and OGB-Arxiv \cite{hu2020open}, are used in our experiments.
Cora, Cite, and Pubmed are three classical network datasets for validating the effectiveness of GNNs.
However, it is recently found that these three datasets sometimes may not effectively validate the predictive power of different graph learning approaches, due to the relatively small data size and data leakage \cite{hu2020open,shchur2018pitfalls}.
Thus, more massive datasets having better data quality have been proposed to evaluate the performance of different approaches \cite{dwivedi2020benchmarking,hu2020open}.
In our experiment, we additionally use CoauthorCS and OGB-Arxiv as testing datasets.
The details of all benchmarking sets can be checked in the appendix.

\textbf{Evaluation and experimental settings}-Two learning tasks, semi-supervised node classification and semi-supervised node clustering are considered in our experiments.
For the training paradigms of both two learning tasks, we closely follow the experimental scenarios established in the related works \cite{hu2020open,kipf2016semi,velivckovic2018graph,yang2016revisiting}.
For the testing phase of different approaches, we use the test splits that are publicly available for classification tasks, and all nodes for clustering tasks.
The effectiveness of all methods is validated through evaluating the classified nodes using $Accuracy$.
In the training stage, we construct the two-layer network structure (i.e., one hidden layer possessed) for all the baselines and different versions of CATs.
In each set of testing data, all approaches are run ten times to obtain the statistically steady performance.
As for other details related to experimental settings, we leave them in the appendix.

\begin{table*}[htbp]
	\centering
	\caption{Average $Accuracy$ on semi-supervised node classification. Bold fonts mean CAT obtains a better performance than any other baseline.}
	\label{classification}
	\begin{tabular}{c|ccccc}
		\hline\hline
		&\bf Cora&\bf Cite&\bf Pubmed&\bf CoauthorCS& \bf OGB-Arxiv\\
		\hline
		MoNet&81.96 $\pm$ 0.50&64.22 $\pm$ 0.16&79.78  $\pm$ 0.33&91.96 $\pm$ 0.75&47.71 $\pm$ 0.27\\
		GCN&81.42 $\pm$ 0.19&71.60 $\pm$ 0.73&79.66 $\pm$ 0.39&91.54 $\pm$ 0.43&71.78 $\pm$ 0.16\\
		GraphSAGE&81.12 $\pm$ 0.41&71.06 $\pm$ 0.64&79.04 $\pm$ 0.62&93.06 $\pm$ 0.80&69.07 $\pm$ 0.27\\
		JKNet&78.34 $\pm$ 0.02& 65.88 $\pm$ 0.01&79.88 $\pm$ 0.01&89.62 $\pm$ 0.01&64.91 $\pm$ 0.01\\
		APPNP&82.80  $\pm$ 0.32& 72.38 $\pm$ 0.50& 82.62 $\pm$ 0.37& 89.16 $\pm$ 0.65& 63.16 $\pm$ 0.54\\
		SGC&81.90  $\pm$ 0.01&71.40  $\pm$ 0.01& 82.42 $\pm$ 0.04& 93.60 $\pm$ 0.01&61.06  $\pm$ 0.09\\
		ARMA& 80.06 $\pm$ 0.57& 70.00 $\pm$ 0.66&76.46  $\pm$ 0.58& 86.28 $\pm$ 0.75& 68.77 $\pm$ 0.17\\
		GIN &81.58 $\pm$ 0.62 &66.90 $\pm$ 0.16 &80.76 $\pm$ 0.33 &93.03 $\pm$ 0.74 &64.02 $\pm$ 0.18\\
		\hline
		GAT&83.84 $\pm$ 0.61& 70.36 $\pm$ 0.42& 81.50 $\pm$ 0.47& 92.80 $\pm$ 0.41& 72.39 $\pm$ 0.07\\
		GAT-$k$-Lap&84.10 $\pm$ 0.24&71.18 $\pm$ 0.52 &82.56 $\pm$ 0.30 &92.70 $\pm$ 0.31&72.47 $\pm$ 0.06\\
		GAT-NetMF&84.44 $\pm$ 0.19 &70.94 $\pm$ 0.16 &81.90 $\pm$ 0.33&93.16 $\pm$ 0.27 &72.42 $\pm$ 0.08\\
		GAT-Deep&83.68 $\pm$ 0.67 & 69.70 $\pm$ 0.57 & 80.13 $\pm$ 0.26 & 92.93 $\pm$ 0.17 & 72.79 $\pm$ 0.09 \\
		\hline
		CAT-I-MF& \bf85.38 $\pm$ 0.16& \bf73.22 $\pm$ 0.19&\bf83.90 $\pm$ 0.24&\bf93.74  $\pm$ 0.14 &\bf72.89 $\pm$ 0.06\\
		CAT-I-SC & \bf85.50 $\pm$ 0.22& \bf73.18 $\pm$ 0.22& \bf84.28 $\pm$ 0.20& \bf93.70 $\pm$ 0.11& \bf72.85 $\pm$ 0.04\\
		CAT-E-MF& \bf85.56 $\pm$ 0.19& \bf73.24 $\pm$ 0.21& \bf83.60 $\pm$ 0.17& 93.40 $\pm$ 0.12& \bf72.81 $\pm$ 0.09\\
		CAT-E-SC& \bf85.40 $\pm$ 0.36& \bf73.02 $\pm$ 0.24& \bf84.02 $\pm$ 0.24& 93.30 $\pm$ 0.11& \bf72.83 $\pm$ 0.11\\
		\hline\hline
	\end{tabular}
\end{table*}

\begin{table*}[htbp]
	\centering
	\caption{Average $Accuracy$ on semi-supervised node clustering. Bold fonts mean CAT obtains a better performance than any other baseline.}
	\label{clustering}
	\begin{tabular}{c|ccccc}
		\hline\hline
		&\bf Cora&\bf Cite&\bf Pubmed&\bf CoauthorCS& \bf OGB-Arxiv\\
		\hline
		MoNet&79.42 $\pm$ 0.86&  63.07 $\pm$ 0.11&79.39  $\pm$ 0.61&  88.75 $\pm$ 0.54 &53.08 $\pm$  0.15\\
		GCN&74.25 $\pm$ 0.13& 63.36 $\pm$ 0.87&  77.83 $\pm$ 0.75& 89.74 $\pm$ 0.53& 75.02 $\pm$ 0.07\\
		GraphSAGE&78.46 $\pm$ 0.56&69.00  $\pm$ 0.17& 79.52 $\pm$ 1.13& 90.16 $\pm$ 0.53& 73.50 $\pm$ 0.13\\
		JKNet&75.95 $\pm$ 0.01&65.12  $\pm$ 0.03& 79.52 $\pm$ 0.01& 86.66 $\pm$ 0.01&71.28 $\pm$ 0.01\\
		APPNP&79.93 $\pm$ 0.82& 70.55 $\pm$ 0.85& 82.81 $\pm$ 0.32& 85.93 $\pm$ 0.39& 69.73 $\pm$ 0.67\\
		SGC&79.38 $\pm$ 0.02&69.71  $\pm$ 0.02& 81.64 $\pm$ 0.01& 90.13 $\pm$ 0.01& 71.09 $\pm$0.37\\
		ARMA&77.70 $\pm$ 0.99& 68.38 $\pm$ 0.87& 77.29 $\pm$ 1.11& 84.72 $\pm$ 0.29&69.24 $\pm$ 0.12\\
		GIN &78.25 $\pm$ 0.46 & 67.83 $\pm$ 0.15 & 79.31 $\pm$ 0.35 & 89.97 $\pm$ 0.26 & 63.85 $\pm$ 0.18\\
		\hline
		GAT&81.39 $\pm$ 0.18& 69.20 $\pm$ 0.28& 80.88 $\pm$ 0.33& 90.09 $\pm$ 0.15& 76.04 $\pm$ 0.38\\
		GAT-$k$-Lap&80.66 $\pm$ 0.31 & 69.56 $\pm$ 0.34 & 81.59 $\pm$ 0.09 & 89.83 $\pm$ 0.18 & 76.21 $\pm$ 0.06 \\
		GAT-NetMF &81.75 $\pm$ 0.26 & 68.96 $\pm$ 0.21 & 81.74 $\pm$ 0.18 & 89.85 $\pm$ 0.21 & 76.06 $\pm$ 0.07\\
		GAT-Deep &81.08 $\pm$ 0.41 & 68.27 $\pm$ 0.06 & 80.55 $\pm$ 0.11 & 89.70 $\pm$ 0.27 & 76.91 $\pm$ 0.15\\
		\hline
		CAT-I-MF& \bf82.17 $\pm$ 0.11& \bf71.15 $\pm$ 0.12& 82.77 $\pm$ 0.07& \bf90.26 $\pm$ 0.22 &\bf77.72 $\pm$ 0.07\\
		CAT-I-SC& \bf82.26 $\pm$ 0.13&\bf71.17 $\pm$ 0.15& \bf 82.86 $\pm$ 0.07& \bf90.29 $\pm$ 0.21& \bf77.01 $\pm$ 0.16\\
		CAT-E-MF&\bf81.98 $\pm$ 0.19& \bf71.21 $\pm$ 0.12&82.40 $\pm$ 0.08& 89.66 $\pm$ 0.22& \bf76.93 $\pm$ 0.16\\
		CAT-E-SC& \bf82.01 $\pm$ 0.24&  \bf71.11 $\pm$ 0.24& 82.61 $\pm$ 0.14& 89.72 $\pm$ 0.15 &\bf76.98 $\pm$ 0.08\\
		\hline\hline
	\end{tabular}
\end{table*}

\subsection{Results on node classification}
The results on semi-supervised node classification are summarized in Table \ref{classification}.
As the table shows, CATs utilizing different attention strategies generally perform better than any other baseline in all the testing datasets.
Specifically, CAT utilizing \textit{Implicit direction} (CAT-I-MF and CAT-I-SC) performs better than all the compared baselines in all the five datasets.
CAT utilizing \textit {Explicit direction} (CAT-E-MF and CAT-E-SC) is better than other compared baselines in four datasets out of five, except the case of CoauthorCS.
In that dataset, CAT-E ranks the second-best when compared with other baselines.

\subsection{Results on node clustering}
Node clustering can be more challenging as all the nodes containing various potential structures in the graph are used in the testing phase.
The results obtained show that CATs still performs robustly when compared with other baselines on this challenging task.
As Table \ref{clustering} shows, the \textit {Implicit direction} strategy utilized by CAT can still ensure the proposed neural architecture to outperform other compared baselines in all the datasets.
As for CAT utilizing \textit {Explicit direction}, it ranks best on three datasets out of five.
While on the remaining two datasets, Pubmed and CoauthorCS, the performance of CAT-E is competitive to the best.
Based on the robust performance shown in Tables \ref{classification} and \ref{clustering}, CAT is observed to be one of the most effective GNNs for various graph learning tasks.

\begin{figure}[!htb]
	\centering
	\begin{subfigure}[b]{0.4\linewidth}
		\includegraphics[width=\textwidth]{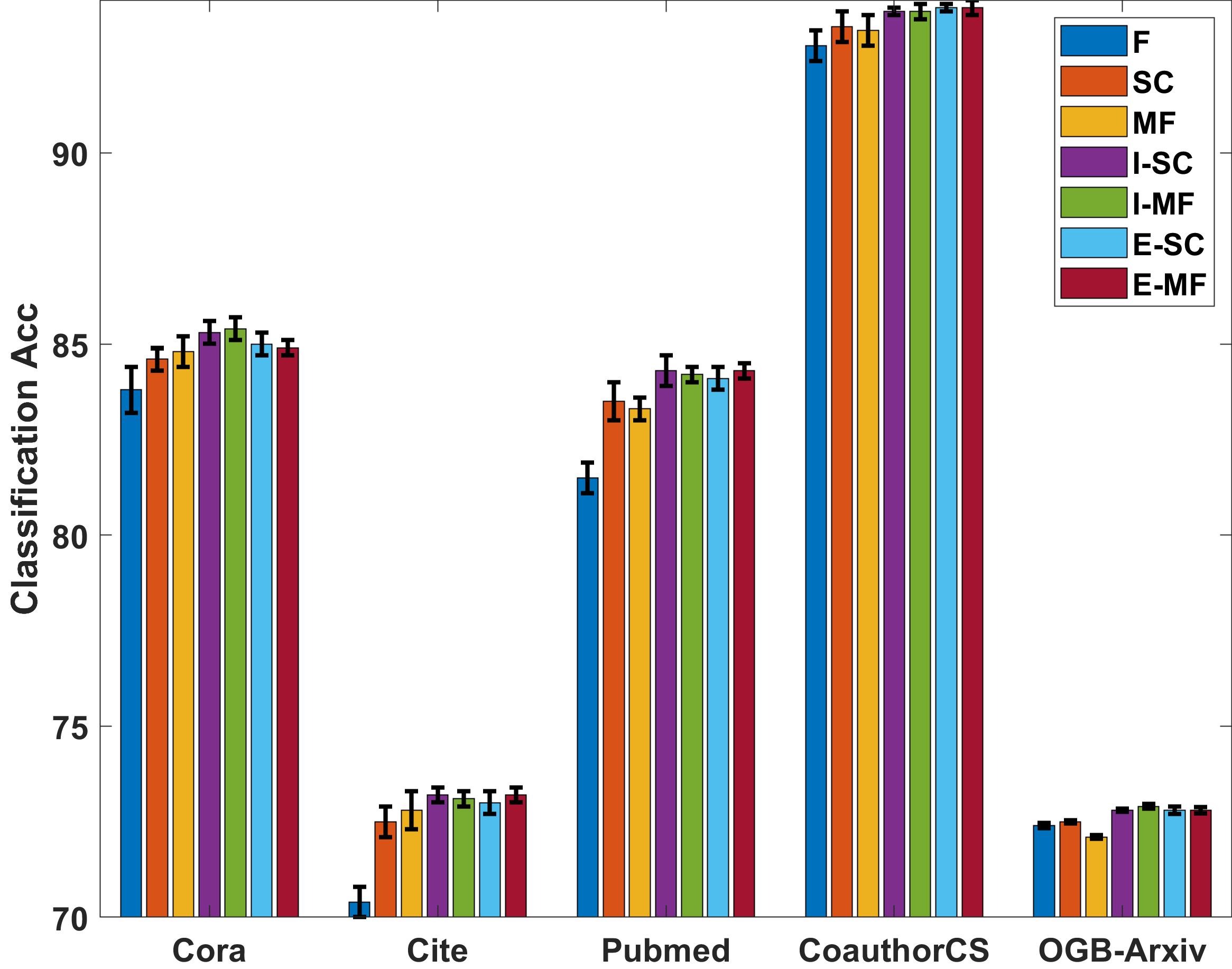}
	\end{subfigure}
	\begin{subfigure}[b]{0.4\linewidth}
		\includegraphics[width=\textwidth]{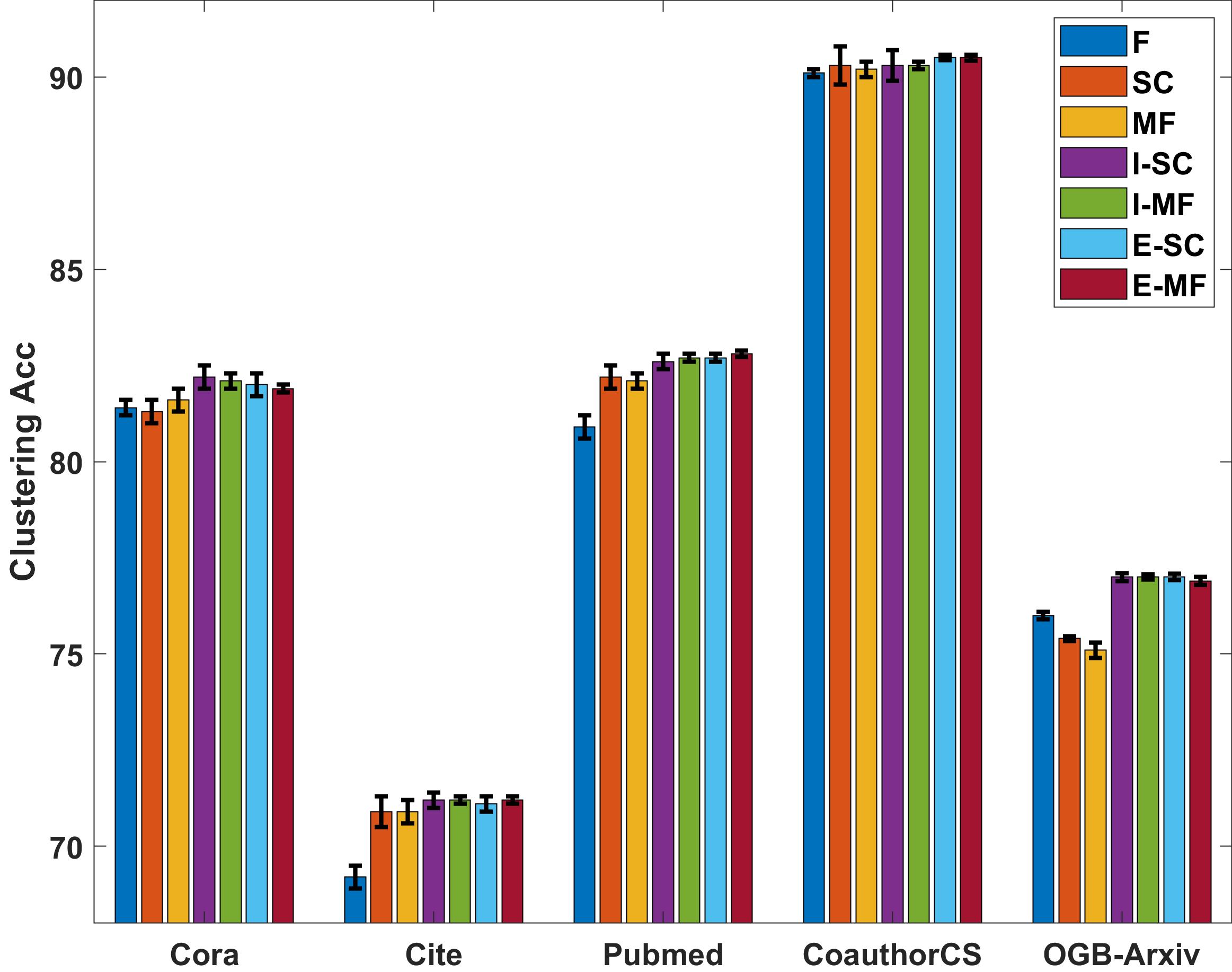}
	\end{subfigure}
	\caption{Performance comparison on computing attention scores using different factors}\label{ablation}
\end{figure}

\subsection{Ablation study}
To further investigate whether the proposed CA mechanisms are effective in improving the predictive power of CATs, we compared the performance of CAs with that obtained by attention mechanisms considering various factors.
Specifically, we let GAT computes attention coefficients using either structural interventions, i.e., node-node correlations pertaining to node cluster embeddings (MF in Eq. (\ref{clustering})) and self-representation coefficients (SC in Eq. (\ref{global})), or node features (Eq.(\ref{f-att}), F).
Then GAT utilizing different attention strategies is used to perform node classification and clustering tasks on all the testing datasets.
The performance comparisons between CATs and GAT utilizing the aforementioned attentions have been summarized in Fig. \ref{ablation}.
As the figure shows, on both classification and clustering tasks, the proposed CA mechanisms perform statistically better than other attention strategies utilized by GAT.
It is also observed that the consideration of structural attentions (SC and MF) may also improve the performance of attention-based GNNs.
Such results thus experimentally agree with and validates our theoretical analysis in Section \ref{theory}.

\section{Discussions}
In this section, some further discussions that may provide better understandings to the proposed attention mechanisms are presented.

\subsection{Comparisons between CATs and GATs with augmented node features}
Besides the proposed Conjoint Attention mechanisms, directly concatenating input node features and structural embeddings \cite{perozzi2014deepwalk,grover2016node2vec,ribeiro2017struc2vec,tang2015line,qiu2018network} is another effective way to make current GNNs more structural bias.
In our experiments, it is shown that the learning performance have improved in most datasets when GAT uses the concatenation of original node features and various structural embeddings.
However, in most datasets, such performance improvement is not as significant as that obtained by CATs. 
Different from directly concatenating the original node input features and structural embeddings for learning node representations, CATs provide a smooth means to compute attention coefficients that jointly consider the diverse relevance regarding to both layer-wise node embeddings and external factors, like the structural interventions pertaining to the correlations generated by the node cluster embeddings considered in this paper.
As a result, CATs can learn node representations from those nodes that are heterogeneously relevant and attain notable learning performance on all the testing datasets.

\subsection{Potential limitations of the proposed approach}
Although the proposed Conjoint Attentions are very effective in improving the learning performance of attention-based GNNs, they may also have possible shortcomings.
First, the predictive power of the proposed CATs may be determined by the quality of the structural interventions.
As the proposed Conjoint Attentions attempt to compute attention scores considering heterogeneous factors, their performance might be negatively affected by those contaminated ones.
However, some potential methods may mitigate the side-effect brought by possible false/noisy external factors. 
One is to utilize adversarial learning modules \cite{lowd2005adversarial} to alleviate the model vulnerability to external contamination. The other method is to consider learning effective factors from multiple data sources (i.e., multi-view).
In previous works \cite{xu2013survey}, multi-view learning has shown to be effective even when some view of data is contaminated.

Second, both space and time complexity of the proposed Conjoint Attentions can be higher than the empirical attention-based GNNs, e.g., GAT.
Thus, how to select a simple but effective strategy for capturing the interventions is crucial for the proposed Conjoint Attentions.
In our experiments, we recorded the memory consumption of CATs when performing different learning tasks in all the testing datasets and the corresponding results are provided in the appendix.
We find that some simple learning paradigms, e.g., matrix factorization (Eq. (\ref{local}) in the manuscript) can ensure CATs to outperform other baselines, while the space and time complexity does not increase much when compared with GATs.
As for the SC strategy, its space complexity is relatively high, if it uses a single-batch optimization method.
Therefore, more efficient optimization methods should be considered when CATs use SC to learn $\mathbf C_{ij}$.

Third, the expressive power of the proposed Conjoint Attention-based GNNs reaches the upper bound of the 1-WL test in the countable feature space, but such discriminative capacity may not be always held by CATs in the uncountable feature space. 
Previous works have proved that the expressive power of a simplex function for feature aggregation in a GNN can surely reach the upper bound of the 1-WL test only in countable feature space, multiple categories of functions for feature aggregations are required to maintain the expressive power of a GNN when the feature space is uncountable \cite{corso2020principal}.
As all Conjoint Attentions belong to one category of function, i.e., mean aggregator, in this paper, we perform the theoretical analysis on the expressive power of CATs assuming the feature space is countable.
Ideally,  the expressive power of CATs can be further improved in uncountable space, if the proposed Conjoint Attentions are appropriately combined with other types of feature aggregators.

\section{Conclusion}\label{conclusion}
In this paper, we have proposed a class of novel attention strategies, known as Conjoint Attentions (CAs) to construct Graph conjoint attention networks (CATs).
Different from empirical graph attentions, CAs offer flexible incorporation of both layer-wise node features and structural interventions that can be learned outside of the GNN to compute appropriate weights for feature aggregation.
Besides, the expressive power of CATs is theoretically validated to reach the upper bound of all message-passing GNNs.
The proposed CATs have been compared with a number of prevalent approaches in different learning tasks.
The obtained notable results verify the CATs' model effectiveness.
In future, we will further improve the effectiveness of CATs in the following ways.
First, besides node cluster embeddings and self-expressiveness, more structural interventions will be explored to compute more compelling attention coefficients for node representation learning.
Second, appropriate adversarial strategies to reduce model vulnerability to contaminated factors that are used for computing attention scores will be considered.
Last but not the least, the proposed conjoint attentions will be extended to learn node representations in multi-view context and heterogeneous graph data.


\begin{ack}
	The authors would like to thank the anonymous reviewers for their constructive comments and suggestions.
	This work is supported in part by the Data Science $\And$ Artificial Intelligence Research Center (DSAIR), Nanyang Technological University, and in part by Agency for Science, Technology and Research (A*STAR).
\end{ack}

\bibliographystyle{plain}

\newpage
\appendix

\section{Proof of Theorem 1}
To prove Theorem 1, we need to consider the two directions of the iff conditions.
If given $c_1 = c_2$, $S_1 = S_2$, and $\sum_{y=x, y\in X_1} f_{c_1y}-\sum_{y=x, y\in X_2} f_{c_2y} = q[\sum_{y=x, y\in X_2} s_{c_2y} - \sum_{y=x, y\in X_1} s_{c_1y}]$, for $q = \frac{r_s}{r_f}$, for the aggregation function utilizing the weights computed by Eq. (7) in the manuscript, we have:
\begin{equation}\label{hx-ip}
	\begin{aligned}
		&h(c_i, X_i) = \sum_{x\in X_i} \alpha_{c_ix}g(x),\\
		&\alpha_{c_ix} = r_f \cdot f_{c_ix} + r_s \cdot s_{c_ix},\\
		& f_{c_ix} = \frac{\exp{ (m_{c_ix})}}{\sum_{x\in X_i}\exp{ (m_{c_ix})}}, s_{c_ix} = \frac{\exp{ (\mathbf C_{c_ix})}}{\sum_{x\in X_i}\exp{ (\mathbf C_{c_ix})}},
	\end{aligned}
\end{equation}
where $m_{c_ix}$ represents the feature similarity between $c_i$ and $x$.
Given Eq. (\ref{hx-ip}), we may directly derive $h(c_1, X_1)$ and $h(c_2, X_2)$:
\begin{equation}
	\begin{aligned}
		&h(c_1, X_1) = \sum_{x\in X_1}\alpha_{c_1x}g(x)=\sum_{x\in X_1}[r_f \cdot f_{c_1x} + r_s \cdot s_{c_1x}]\cdot g(x)\\
		&h(c_2, X_2) = \sum_{x\in X_2}\alpha_{c_2x}g(x)=\sum_{x\in X_2}[r_f \cdot f_{c_2x} + r_s \cdot s_{c_2x}]\cdot g(x)
	\end{aligned}
\end{equation}
Considering $c_1 = c_2$, $S_1 = S_2$, and $\sum_{y=x, y\in X_1} f_{c_1y}-\sum_{y=x, y\in X_2} f_{c_2y} = q[\sum_{y=x, y\in X_2} s_{c_2y} - \sum_{y=x, y\in X_1} s_{c_1y}]$, for $q = \frac{r_s}{r_f}$, we directly derive $h(c_1, X_1)=h(c_2, X_2)$.

If we are given $h(c_1, X_1) = h(c_2, X_2)$, we are able to prove that the conditions mentioned in the theorem are necessary by showing contradictions occur when they are not satisfied.
If $h(c_1, X_1) = h(c_2, X_2)$, we have:
\begin{equation}\label{h1minush2-ip}
	\begin{aligned}
		&h(c_1, X_1)-h(c_2, X_2) = \\
		&\sum_{x\in X_1}[r_f \cdot f_{c_1x} + r_s \cdot s_{c_1x}]\cdot g(x) - \sum_{x\in X_2}[r_f \cdot f_{c_2x} + r_s \cdot s_{c_2x}]\cdot g(x)=0
	\end{aligned}
\end{equation}
Firstly, assuming $S_1 \ne S_2$, for any $g(\cdot)$, we thereby have:
\begin{equation}\label{s1nes2-ip}
	\begin{aligned}
		&h(c_1, X_1)-h(c_2, X_2) = \sum_{x\in S_1 \cap S_2}\![\sum_{y=x, y\in X_1}[r_f \cdot f_{c_1y} + r_s \cdot s_{c_1y}]\\
		&- \sum_{y=x, y\in X_2}[r_f \cdot f_{c_2y} + r_s \cdot s_{c_2y}]]\cdot g(x)\\
		&+\sum_{x\in S_1 \setminus S_2}\!\sum_{y=x, y\in X_1}[r_f \cdot f_{c_1y} + r_s \cdot s_{c_1y}]\cdot g(x)\\
		&-\sum_{x\in S_2 \setminus S_1}\sum_{y=x, y\in X_2}[r_f \cdot f_{c_2y} + r_s \cdot s_{c_2y}]\cdot g(x)=0
	\end{aligned}
\end{equation}
As Eq. (\ref{s1nes2-ip}) holds for any $g(\cdot)$, we may define another function $g^\prime(\cdot)$ as follows:
\begin{equation}\label{gprime-ip}
	\begin{aligned}
		&g(x) = g^\prime(x), \text {for } x\in S_1\cap S_2\\
		&g(x) = g^\prime(x) - 1, \text {for }x \in S_1\setminus S_2\\
		&g(x) = g^\prime(x) + 1, \text {for } x\in S_2\setminus S_1\\
	\end{aligned}
\end{equation}
If Eq. (\ref{s1nes2-ip}) holds, we also have:
\begin{equation}\label{s1nes3-ip}
	\begin{aligned}
		&h(c_1, X_1)-h(c_2, X_2) = \sum_{x\in S_1 \cap S_2}\![\sum_{y=x, y\in X_1}[r_f \cdot f_{c_1y} + r_s \cdot s_{c_1y}]\\
		&- \sum_{y=x, y\in X_2}[r_f \cdot f_{c_2y} + r_s \cdot s_{c_2y}]]\cdot g^\prime(x)\\
		&+\sum_{x\in S_1 \setminus S_2}\!\sum_{y=x, y\in X_1}[r_f \cdot f_{c_1y} + r_s \cdot s_{c_1y}]\cdot g^\prime(x)\\
		&-\sum_{x\in S_2 \setminus S_1}\sum_{y=x, y\in X_2}[r_f \cdot f_{c_2y} + r_s \cdot s_{c_2y}]\cdot g^\prime(x)\\
		&=\sum_{x\in S_1 \cap S_2}\![\sum_{y=x, y\in X_1}[r_f \cdot f_{c_1y} + r_s \cdot s_{c_1y}]\\
		&- \sum_{y=x, y\in X_2}[r_f \cdot f_{c_2y} + r_s \cdot s_{c_2y}]]\cdot g(x)\\
		&+\sum_{x\in S_1 \setminus S_2}\!\sum_{y=x, y\in X_1}[r_f \cdot f_{c_1y} + r_s \cdot s_{c_1y}]\cdot [g(x)+1]\\
		&-\sum_{x\in S_2 \setminus S_1}\sum_{y=x, y\in X_2}[r_f \cdot f_{c_2y} + r_s \cdot s_{c_2y}]\cdot [g(x)-1]=0
	\end{aligned}
\end{equation}
As Eq. (\ref{s1nes2-ip}) equals Eq. (\ref{s1nes3-ip}), we have:
\begin{equation}
	\begin{aligned}
		&\sum_{x\in S_1 \setminus S_2}\!\sum_{y=x, y\in X_1}[r_f \cdot f_{c_1y} + r_s \cdot s_{c_1y}] + \sum_{x\in S_2 \setminus S_1}\sum_{y=x, y\in X_2}[r_f \cdot f_{c_2y} + r_s \cdot s_{c_2y}]=0
	\end{aligned}
\end{equation}
Obviously, the above equation does not hold as the terms in the summation operator are positive.
Thus, $S_1 \ne S_2$ is not true.
We may now assume $S_1=S_2=S$. Eliminating the irrational terms in Eq. (\ref{s1nes2-ip}), we have:
\begin{equation}
	\begin{aligned}
		&\sum_{x\in S_1 \cap S_2}\![\sum_{y=x, y\in X_1}[r_f \cdot f_{c_1y} + r_s \cdot s_{c_1y}]- \sum_{y=x, y\in X_2}[r_f \cdot f_{c_2y} + r_s \cdot s_{c_2y}]]\cdot g(x)=0
	\end{aligned}
\end{equation}
Thus we know each term in the summation equals zero:
\begin{equation}\label{s1es2-ip}
	\begin{aligned}
		&\sum_{y=x, y\in X_1}[r_f \cdot f_{c_1y} + r_s \cdot s_{c_1y}]- \sum_{y=x, y\in X_2}[r_f \cdot f_{c_2y} + r_s \cdot s_{c_2y}]=0
	\end{aligned}
\end{equation}
As external factors and node feature similarity are heterogeneous, we may assume $\sum_{y=x, y\in X_1} r_s \cdot s_{c_1y} = \sum_{y=x, y\in X_2} r_s \cdot s_{c_2y}$.
Eq. (\ref{s1es2-ip}) can be simplified and rewritten as:
\begin{equation}\label{s1es22-ip}
	\begin{aligned}
		&\frac{\mu_1(x)}{\mu_2(x)}=\frac{\exp{ (m_{c_2x})}\sum_{x\in S_1}\!\sum_{y=x, y\in X_1}\!\exp{ (m_{c_1x})}}{\exp{ (m_{c_1x})}\sum_{x\in S_2}\sum_{y=x, y\in X_2}\exp{ (m_{c_2x})}}
	\end{aligned}
\end{equation}
It is obvious that LHS of Eq. (\ref{s1es22-ip}) is a rational number.
However, the RHS of Eq. (\ref{s1es22-ip}) can be an irrational number.
We may consider $S = \lbrace s, s_0\rbrace$ and assume $c_1 = s_0$, $c_2 = s$. We may also assume the feature similarity between central node and others as follows:
\begin{equation}
	\begin{aligned}
		&m_{c_1x} = 1, \text{for } x \in S\\
		&m_{c_2s} = 1, m_{c_2s_0} = 2
	\end{aligned}
\end{equation}
Consider $x = s$, we have:
\begin{equation}\label{nemu}
	\begin{aligned}
		&\frac{\mu_1(s)}{\mu_2(s)}=\frac{\vert X_1\vert}{\vert X_2\vert - n + ne},
	\end{aligned}
\end{equation}
where $n$ stands for the number of $s_0$ in $X_2$. It is obvious that the above equality does not hold as the RHS is an irrational number, while LHS is a rational number.
Thus $c_1 \ne c_2$ is false.
Given $c_1 = c_2$, Eq. (\ref{s1es2-ip}) can be rewritten as:
\begin{equation}\label{s1es2-ipre}
	\begin{aligned}
		&\sum_{y=x, y\in X_1}r_f \cdot f_{c_1y}-\sum_{y=x, y\in X_2} r_f \cdot f_{c_2y} + \sum_{y=x, y\in X_1} r_s \cdot s_{c_1y}- \sum_{y=x, y\in X_2} r_s \cdot s_{c_2y}=0
	\end{aligned}
\end{equation}
To ensure above equation holds, we have:
\begin{equation}
	\sum_{y=x, y\in X_1} f_{c_1y}-\sum_{y=x, y\in X_2} f_{c_2y} = \frac{r_s}{r_f}[\sum_{y=x, y\in X_2} s_{c_2y} - \sum_{y=x, y\in X_1} s_{c_1y}]
\end{equation}
Further denoting $\frac{r_s}{r_f} = q$, we have $\sum_{y=x, y\in X_1} f_{c_1y}-\sum_{y=x, y\in X_2} f_{c_2y} = q[\sum_{y=x, y\in X_2} s_{c_2y} - \sum_{y=x, y\in X_1} s_{c_1y}]$. $\hfill\square$

\section{Proof of Theorem 2}
To prove Theorem 2, we can follow the procedure which is used to prove Theorem 1.
If given $c_1 = c_2$, $S_1 = S_2$, and $q\cdot \sum_{y = x, y\in X_1}\phi(\mathbf C_{c_1y}) = \sum_{y=x, y\in X_1} \phi(\mathbf C_{c_2y})$, for $q > 0$, we may directly replace $\phi(\cdot)$ using $\exp\lbrace\cdot\rbrace$.
For the aggregation function solely using the weights obtained by Eq. (8) in the manuscript, we have:
\begin{equation}\label{hx}
	\begin{aligned}
		&h(c_i, X_i) = \sum_{x\in X_i} \alpha_{c_ix}g(x),\\
		&\alpha_{c_ix} = \frac{f_{c_ix}\cdot s_{c_ix}}{ \sum_{x\in X_i} f_{c_ix}\cdot s_{c_ix}},\\
		& f_{c_ix} = \frac{\exp{ (m_{c_ix})}}{\sum_{x\in X_i}\exp{ (m_{c_ix})}}, s_{c_ix} = \frac{\exp{ (\mathbf C_{c_ix})}}{\sum_{x\in X_i}\exp{ (\mathbf C_{c_ix})}},
	\end{aligned}
\end{equation}
where $m_{c_ix}$ represents the feature similarity between $c_i$ and $x$.
Given Eq. (\ref{hx}), we may directly derive $h(c_1, X_1)$ and $h(c_2, X_2)$:
\begin{equation}
	\begin{aligned}
		&h(c_1, X_1) = \sum_{x\in X_1}\alpha_{c_1x}g(x)=\sum_{x\in X_1}[\frac{f_{c_1x}\cdot s_{c_1x}}{ \sum_{x\in X_1} f_{c_1x}\cdot s_{c_1x}}]\cdot g(x)\\
		&h(c_2, X_2) = \sum_{x\in X_2}\alpha_{c_2x}g(x)=\sum_{x\in X_2}[\frac{f_{c_2x}\cdot s_{c_2x}}{ \sum_{x\in X_2} f_{c_2x}\cdot s_{c_2x}}]\cdot g(x)
	\end{aligned}
\end{equation}
Given $S_1 = S_2$ and $c_1 = c_2$, $h(c_2, X_2)$ can be rewritten as:
\begin{equation}\label{rewritten-hx2}
	\begin{aligned}
		&h(c_2, X_2) = \sum_{x\in S_2}[\frac{f_{c_2x}\cdot \sum_{y=x, y\in X_2}s_{c_2y}}{ \sum_{x\in S_2} f_{c_2x}\cdot\sum_{y=x, y\in X_2}s_{c_2y}}]\cdot g(x)\\
		&=\sum_{x\in S_2}[\frac{\frac{\exp{ (m_{c_2x})}}{\sum_{y\in X_2}\exp{ (m_{c_2y})}}\cdot \sum_{y=x, y\in X_2}\frac{\exp{ (\mathbf C_{c_2y})}}{\sum_{y\in X_2}\exp{ (\mathbf C_{c_2y})}}}{ \sum_{x\in S_2} \frac{\exp{ (m_{c_2x})}}{\sum_{y\in X_2}\exp{ (m_{c_2y})}}\sum_{y=x, y\in X_2}\frac{\exp{ (\mathbf C_{c_2y})}}{\sum_{y\in X_2}\exp{ (\mathbf C_{c_2y})}}}]\cdot g(x)\\
		&=\sum_{x\in S_2}\frac{\exp{ (m_{c_2x})}\sum_{y=x, y\in X_2}\exp{ (\mathbf C_{c_2y})}}{\sum_{x\in S_2}\exp{ (m_{c_2x})}\sum_{y=x, y\in X_2}\exp{ (\mathbf C_{c_2y})}}\cdot g(x)
	\end{aligned}
\end{equation}
Given $q\cdot \sum_{x\in X_1} \exp\lbrace\mathbf C_{c_1x}\rbrace = \sum_{y=x} \exp\lbrace\mathbf C_{c_2y}\rbrace$, Eq. (\ref{rewritten-hx2}) is equivalent to:
\begin{equation}\label{rewritten-hx3}
	h(c_2, X_2) = \sum_{x\in S_1}\frac{q\cdot \exp{ (m_{c_2x})}\sum_{ y=x, y\in X_1}\exp{ (\mathbf C_{c_1y})}}{\sum_{x\in S_1}q\cdot\exp{ (m_{c_2x})}\sum_{y=x, y\in X_1}\exp{ (\mathbf C_{c_1y})}}\cdot g(x)
\end{equation}
Considering $c_1 = c_2$, we have $h(c_1, X_1) = h(c_2, X_2)$.

If we are given $h(c_1, X_1) = h(c_2, X_2)$, we are able to prove the conditions mentioned in the theorem are necessary by showing contradictions occur when they are not satisfied.
If $h(c_1, X_1) = h(c_2, X_2)$, we have:
\begin{equation}\label{h1minush2}
	\begin{aligned}
		&h(c_1, X_1)-h(c_2, X_2) = \\
		&\sum_{x\in X_1}[\frac{f_{c_1x}\cdot s_{c_1x}}{ \sum_{x\in X_1} f_{c_1x}\cdot s_{c_1x}}]\cdot g(x) - \sum_{x\in X_2}[\frac{f_{c_2x}\cdot s_{c_2x}}{ \sum_{x\in X_2} f_{c_2x}\cdot s_{c_2x}}]\cdot g(x)=0
	\end{aligned}
\end{equation}
Firstly, assuming $S_1 \ne S_2$, for any $g(\cdot)$, we thereby have:
\begin{equation}\label{s1nes2}
	\begin{aligned}
		&h(c_1, X_1)-h(c_2, X_2) = \sum_{x\in S_1 \cap S_2}\![\frac{\exp{ (m_{c_1x})}\sum_{y=x, y\in X_1}\!\exp{ (\mathbf C_{c_1y})}}{\sum_{x\in S_1}\!\exp{ (m_{c_1x})}\sum_{y=x, y\in X_1}\!\exp{ (\mathbf C_{c_1y})}}\\
		&- \frac{\exp{ (m_{c_2x})}\sum_{y=x, y\in X_2}\!\exp{ (\mathbf C_{c_2y})}}{\sum_{x\in S_2}\!\exp{ (m_{c_2x})}\sum_{y=x, y\in X_2}\!\exp{ (\mathbf C_{c_2y})}}]\cdot g(x)\\
		&+\sum_{x\in S_1 \setminus S_2}\![\frac{\exp{ (m_{c_1x})}\sum_{y=x, y\in X_1}\!\exp{ (\mathbf C_{c_1y})}}{\sum_{x\in S_1}\!\exp{ (m_{c_1x})}\sum_{y=x, y\in X_1}\!\exp{ (\mathbf C_{c_1y})}}]\cdot g(x)\\
		&-\sum_{x\in S_2 \setminus S_1}[\frac{\exp{ (m_{c_2x})}\sum_{y=x, y\in X_2}\!\exp{ (\mathbf C_{c_2y})}}{\sum_{x\in S_2}\!\exp{ (m_{c_2x})}\sum_{y=x, y\in X_2}\!\exp{ (\mathbf C_{c_2y})}}]\cdot g(x)=0
	\end{aligned}
\end{equation}
As Eq. (\ref{s1nes2}) holds for any $g(\cdot)$, we may define another function $g^\prime(\cdot)$ as shown in Eq. (\ref{gprime-ip}).
If Eq. (\ref{s1nes2}) holds, we also have:
\begin{equation}\label{s1nes3}
	\begin{aligned}
		&h(c_1, X_1)-h(c_2, X_2) = \sum_{x\in S_1 \cap S_2}\![\frac{\exp{ (m_{c_1x})}\sum_{y=x, y\in X_1}\!\exp{ (\mathbf C_{c_1y})}}{\sum_{x\in S_1}\!\exp{ (m_{c_1x})}\sum_{y=x, y\in X_1}\!\exp{ (\mathbf C_{c_1y})}}\\
		&- \frac{\exp{ (m_{c_2x})}\sum_{y=x, y\in X_2}\!\exp{ (\mathbf C_{c_2y})}}{\sum_{x\in S_2}\!\exp{ (m_{c_2x})}\sum_{y=x, y\in X_2}\!\exp{ (\mathbf C_{c_2y})}}]\cdot g^\prime(x)\\
		&+\sum_{x\in S_1 \setminus S_2}\![\frac{\exp{ (m_{c_1x})}\sum_{y=x, y\in X_1}\!\exp{ (\mathbf C_{c_1y})}}{\sum_{x\in S_1}\!\exp{ (m_{c_1x})}\sum_{y=x, y\in X_1}\!\exp{ (\mathbf C_{c_1y})}}]\cdot g^\prime(x)\\
		&-\sum_{x\in S_2 \setminus S_1}[\frac{\exp{ (m_{c_2x})}\sum_{y=x, y\in X_2}\!\exp{ (\mathbf C_{c_2y})}}{\sum_{x\in S_2}\!\exp{ (m_{c_2x})}\sum_{y=x, y\in X_2}\!\exp{ (\mathbf C_{c_2y})}}]\cdot g^\prime(x)\\
		&=\sum_{x\in S_1 \cap S_2}\![\frac{\exp{ (m_{c_1x})}\sum_{y=x, y\in X_1}\!\exp{ (\mathbf C_{c_1y})}}{\sum_{x\in S_1}\!\exp{ (m_{c_1x})}\sum_{y=x, y\in X_1}\!\exp{ (\mathbf C_{c_1y})}}\\
		&- \frac{\exp{ (m_{c_2x})}\sum_{y=x, y\in X_2}\!\exp{ (\mathbf C_{c_2y})}}{\sum_{x\in S_2}\!\exp{ (m_{c_2x})}\sum_{y=x, y\in X_2}\!\exp{ (\mathbf C_{c_2y})}}]\cdot g(x)\\
		&+\sum_{x\in S_1 \setminus S_2}\![\frac{\exp{ (m_{c_1x})}\sum_{y=x, y\in X_1}\!\exp{ (\mathbf C_{c_1y})}}{\sum_{x\in S_1}\!\exp{ (m_{c_1x})}\sum_{y=x, y\in X_1}\!\exp{ (\mathbf C_{c_1y})}}]\cdot [g(x)+1]\\
		&-\sum_{x\in S_2 \setminus S_1}[\frac{\exp{ (m_{c_2x})}\sum_{y=x, y\in X_2}\!\exp{ (\mathbf C_{c_2y})}}{\sum_{x\in S_2}\!\exp{ (m_{c_2x})}\sum_{y=x, y\in X_2}\!\exp{ (\mathbf C_{c_2y})}}]\cdot [g(x)-1]=0
	\end{aligned}
\end{equation}
As Eq. (\ref{s1nes2}) equals Eq. (\ref{s1nes3}), we have:
\begin{equation}
	\begin{aligned}
		&\sum_{x\in S_1 \setminus S_2}\![\frac{\exp{ (m_{c_1x})}\sum_{y=x, y\in X_1}\!\exp{ (\mathbf C_{c_1y})}}{\sum_{x\in S_1}\!\exp{ (m_{c_1x})}\sum_{y=x, y\in X_1}\!\exp{ (\mathbf C_{c_1y})}}]\\
		& + \sum_{x\in S_2 \setminus S_1}[\frac{\exp{ (m_{c_2x})}\sum_{y=x, y\in X_2}\!\exp{ (\mathbf C_{c_2y})}}{\sum_{x\in S_2}\!\exp{ (m_{c_2x})}\sum_{y=x, y\in X_2}\!\exp{ (\mathbf C_{c_2y})}}]=0
	\end{aligned}
\end{equation}
Obviously, the above equation does not hold as softmax function is positive.
Thus, $S_1 \ne S_2$ is not true.
We may now assume $S_1=S_2=S$. Eliminating the irrational terms in Eq. (\ref{s1nes2}), we have:
\begin{equation}
	\begin{aligned}
		&\sum_{x\in S_1 \cap S_2}\![\frac{\exp{ (m_{c_1x})}\sum_{y=x, y\in X_1}\!\exp{ (\mathbf C_{c_1y})}}{\sum_{x\in S_1}\!\exp{ (m_{c_1x})}\sum_{y=x, y\in X_1}\!\exp{ (\mathbf C_{c_1y})}}\\
		&- \frac{\exp{ (m_{c_2x})}\sum_{y=x, y\in X_2}\!\exp{ (\mathbf C_{c_2y})}}{\sum_{x\in S_2}\!\exp{ (m_{c_2x})}\sum_{y=x, y\in X_2}\!\exp{ (\mathbf C_{c_2y})}}]\cdot g(x)=0
	\end{aligned}
\end{equation}
Thus we know each term in the summation equals zero:
\begin{equation}\label{s1es2}
	\begin{aligned}
		&\frac{\exp{ (m_{c_1x})}\sum_{y=x, y\in X_1}\exp{ (\mathbf C_{c_1y})}}{\sum_{x\in S_1}\exp{ (m_{c_1x})}\sum_{y=x, y\in X_1}\exp{ (\mathbf C_{c_1y})}}\\
		&- \frac{\exp{ (m_{c_2x})}\sum_{y=x, y\in X_2}\exp{ (\mathbf C_{c_2y})}}{\sum_{x\in S_2}\exp{ (m_{c_2x})}\sum_{y=x, y\in X_2}\exp{ (\mathbf C_{c_2y})}}=0
	\end{aligned}
\end{equation}
Eq. (\ref{s1es2}) is equivalent to:
\begin{equation}\label{s1es22}
	\begin{aligned}
		&\frac{\sum_{y=x, y\in X_1}\!\exp{ (\mathbf C_{c_1y})}}{\sum_{y=x, y\in X_2}\!\exp{ (\mathbf C_{c_2y})}}=\frac{\exp{ (m_{c_2x})}\sum_{x\in S_1}\!\exp{ (m_{c_1x})}\sum_{y=x, y\in X_1}\!\exp{ (\mathbf C_{c_1y})}}{\exp{ (m_{c_1x})}\sum_{x\in S_2}\!\exp{ (m_{c_2x})}\!\sum_{y=x, y\in X_2}\!\exp{ (\mathbf C_{c_2y})}}
	\end{aligned}
\end{equation}
We may consider $S = \lbrace s, s_0\rbrace$ and assume $c_1 = s_0$, $c_2 = s$. We may also assume the feature similarity between central node and others as follows:
\begin{equation}
	\begin{aligned}
		&m_{c_1x} = 1, \text{for } x \in S\\
		&m_{c_2s} = 1, m_{c_2s_0} = 2
	\end{aligned}
\end{equation}
Consider $x = s$, we have:
\begin{equation}
	\begin{aligned}
		&\frac{\sum_{s\in X_1}\exp{ (\mathbf C_{c_1s})}}{\sum_{s\in X_2}\exp{ (\mathbf C_{c_2s})}}=\frac{e [e\sum_{s\in X_1}\exp{ (\mathbf C_{c_1s})} + e\sum_{s_0\in X_1}\exp{ (\mathbf C_{c_1s_0})}]}{e[e\sum_{s\in X_2}\exp{ (\mathbf C_{c_2s})} + e^2\sum_{s_0\in X_2}\exp{ (\mathbf C_{c_2s_0})}]}
	\end{aligned}
\end{equation}
As the learning of $\mathbf C$ is independent of feature mapping, and the computation of attention coefficients, $\exp{ (\mathbf C_{cx})}$ can be any positive value.
By setting the exponential values in the above equation as $a$, which is a positive value. We have $\frac{\mu_1(s)}{\mu_2(s)}=\frac{|X_1|}{|X_2|-n+ne}$. Similar with Eq. (\ref{nemu}), $c_1 \ne c_2$ is not true.
Since $c_1 =  c_2 = c$, Eq. (\ref{s1es22}) can be rewritten as:
\begin{equation}\label{c1ec2}
	\begin{aligned}
		&\frac{\sum_{y=x, y\in X_1}\!\exp{ (\mathbf C_{cy})}}{\sum_{y=x, y\in X_2}\!\exp{ (\mathbf C_{cy})}}=\frac{\sum_{x\in S_1}\!\exp{ (m_{cx})}\sum_{y=x, y\in X_1}\!\exp{ (\mathbf C_{cy})}}{\sum_{x\in S_2}\!\exp{ (m_{cx})}\!\sum_{y=x, y\in X_2}\!\exp{ (\mathbf C_{cy})}}=const > 0.
	\end{aligned}
\end{equation}
By setting $const$ as $\frac{1}{q}$ and $\exp{ (\mathbf C_{cy})} = \phi(\mathbf C_{cy})$ , we have $q\sum_{y=x, y\in X_1}\!\phi(\mathbf C_{cy})=\sum_{y=x, y\in X_2}\!\phi(\mathbf C_{cy})$.$\hfill\square$

\section{Proof of Corollary 1}
According to Theorem 1, we denote $X_1 = (S, \mu_1)$, $X_2 = (S, \mu_2)$, $c \in S$.
We also assume $\sum_{y=x, y\in X_1} f_{c_1y}-\sum_{y=x, y\in X_2} f_{c_2y} = q[\sum_{y=x, y\in X_2} s_{c_2y} - \sum_{y=x, y\in X_1} s_{c_1y}]$, for $q = \frac{r_s}{r_f}$.
When $\mathcal T$ uses the weights obtained solely by Eq. (7) in the manuscript to aggregate node features, it is easy to verify $\sum_{x\in X_1} \alpha_{cx} f(x) = \sum_{x\in X_2} \alpha_{cx} f(x)$, i.e., $\mathcal T$ cannot distinguish the structures satisfied the aforementioned conditions.
When $\mathcal T$ uses the Conjoint Attentions shown in Eq. (9) in the manuscript for feature aggregation and the corresponding attention scores are obtained by the \textit{Implicit Strategy} (Eq. (7) in the manuscript), we have $\sum_{x\in X_1} \alpha_{cx} f(x) - \sum_{x\in X_2} \alpha_{cx} f(x) = \epsilon (\frac{1}{|X_1|}-\frac{1}{|X_2|})\alpha_{cc} f(c)$, where $|X_1| = |\mathcal N_1|$, and $|X_2| = |\mathcal N_2|$.
Since $|X_1| \ne |X_2|$, $\sum_{x\in X_1} \alpha_{cx} f(x) - \sum_{x\in X_2} \alpha_{cx} f(x) \ne 0$, meaning the aggregation function $\mathcal T$ that is based on Eqs. (7) and (9) in the manuscript, can successfully distinguish all the structures that cannot be discriminated by $\mathcal T$ solely based on Eq. (7) in the manuscript.
Similarly, when the proposed Conjoint Attention (Eq. (9) in the manuscript) utilizing the \textit{Explicit Strategy} (Eq. (8) in the manuscript), we can prove such aggregation function also can distinguish those distinct multisets that cannot  be discriminated by the one solely based on the \textit{Explicit Strategy}. $\hfill\square$

\begin{table*}[t]
	\centering
	\caption{Characteristics of the testing datasets used in our experiments}
	\label{data}
	\begin{tabular}{c|ccccc}
		\hline\hline
		& \textbf{Cora} & \textbf{Cite} & \textbf{Pubmed} & \textbf{CoauthorCS}&\textbf{OGB-Arxiv} \\ \hline
		$N$                & 2708          & 3327              & 19717           &18333& 169343              \\
		$\vert E\vert$              & 5429          & 4732              & 44338          &327576 & 1166243             \\
		$D$                & 1433          & 3703              & 500             &6805& 128                 \\
		$C$                & 7             & 6                 & 3               &15& 40                  \\
		\textbf{Training Nodes}   & 140           & 120               & 60       &300       & 90941               \\
		\textbf{Validation Nodes} & 500           & 500               & 500       &500      & 29799               \\
		\textbf{Test Nodes}       & 1000$\backslash$2708         & 1000$\backslash$3327              & 1000$\backslash$19717         &1000$\backslash$18333   & 48603$\backslash$169343               \\ \hline\hline
	\end{tabular}
\end{table*}

\section{More details on the experiments}
In this section, how the experiments used to validate the effectiveness of the proposed Graph conjoint attention networks (CATs) utilizing different Conjoint Attentions are set up is introduced with more details.

\subsection{Dataset description}
Five network datasets, which are Cora, Cite, Pubmed \cite{lu2003link,sen2008collective}, CoauthorCS \cite{shchur2018pitfalls}, and OGB-Arxiv \cite{hu2020open}, are used in our experiments to validate the effectiveness of different approaches.
In Cora, Cite, Pubmed, and OGB-Arxiv, vertices, edges, and vertex features represent the documents, citations between pairwise documents, and the bag-of-words representations of the documents, respectively.
While, in CoauthorCS, the authors, author-author collaborations, and the keywords of the collaborated papers are respectively represented as nodes, edges, and node features.
The statistics of these benchmarking datasets are summarized in Table \ref{data}, where $N$, $\vert E\vert$, $D$, and $C$ denote the number of vertices, edges, vertex features, and the number of classes in each dataset, respectively.

\subsection{Learning tasks for evaluation}
Two learning tasks, that are semi-supervised node classification and semi-supervised node clustering (community detection), are considered in our experiments to validate the effectiveness of different approaches.
To set up the experiments, we closely follow the paradigms used in previous works \cite{hu2020open,velivckovic2018graph,yang2016revisiting}.
All the datasets are split into three parts: training, validation, and testing.
In both two learning tasks, we use the same split of training set in each dataset.
For node classification tasks, a fraction of nodes are used for testing in each dataset, but all are considered in node clustering tasks.
Specifically, in Cora, Cite, Pubmed, and CoauthorCS, only 20 labeled nodes per class are used for training, but all the feature vectors.
As for testing, we use 1000 nodes in each dataset to build the testing split for node classification task.
For dataset OGB-Arxiv, we use a practical split strategy provided in \cite{hu2020open}, segmenting the nodes which represent the academic papers, according to the publication years.
Papers published up to 2017 are in the training split.
While those published in 2018 and 2019 are used for validation and test for semi-supervised node classification, respectively. 

The reason that we use the aforementioned two learning tasks in our experiments is they may test the effectiveness of different approaches in a complementary manner.
Semi-supervised node classification may reflect the predictive accuracy of a learner from a local perspective, as a fraction of nodes in each dataset are used in the testing phase.
In contrast, semi-supervised node clustering may indicate the overall learning performance of an approach, 
when a small number of node labels are used in the training stage.
As all nodes are considered, semi-supervised node clustering may involve more potential structures in the graph, so that the testing stage is more challenging and the power of different graph learning methods can be comprehensively evaluated.

\subsection{Detailed settings of the graph neural networks}
The proposed CATs utilizing different Conjoint Attentions have been compared with a number of prevalent baselines, including Arma filter GNN (ARMA) \citep{bianchi2021graph}, Simplified graph convolutional Networks (SGC) \citep{wu2019simplifying}, Personalized Pagerank GNN (APPNP) \citep{klicperapredict}, Graph attention networks (GAT) \citep{velivckovic2018graph}, Jumping knowledge networks (JKNet) \citep{xu2018representation}, Graph convolutional networks (GCN) \citep{kipf2016semi}, GraphSAGE \citep{hamilton2017inductive}, Mixture model CNN (MoNet) \citep{monti2017geometric}, and Graph isomorphism network (GIN) \cite{xu2018powerful}.
Besides, we also construct several variants of GAT which use the concatenation of original node features and structural embeddings \cite{qiu2018network,perozzi2014deepwalk} to learn representations in all the testing datasets.
To perform an unbiased comparison, we use the publicly released source codes to implement all the selected baselines.
All the compared baselines use a two-layer network structure, i.e. the output layer followed by only one hidden layer, to learn representations for the downstream tasks.
As for the tunable parameters of each baseline, we either use the default settings, or attempt to find the ones that lead the baseline to learn the best representations in each dataset.
Here, we summarize the configurations of pivotal settings of all the compared baselines in Table \ref{exp-setting}, where lr and hidden stand for the learning rate and the dimension of hidden layer, respectively.
As for the settings of CATs, we generally configure them as what we have done with GAT (see Table \ref{exp-setting}).
Specifically, the maximum number of training epochs is set to 1500.
The learning rate in OGB-Arxiv is 0.002, and that in the rest of the datasets is set to 0.01.
All graph neural networks are initialized using Glorot initialization and trained to minimize cross-entropy on the training nodes using the Adam SGD optimizer \cite{kingma2014adam}.
And all the neural networks are trained on a single graphics card, NVIDIA RTX 3090 with 24GB, and are done in the following software environment: Python 3.8, PyTorch 1.8.1, and CUDA 11.1.

\begin{table}[]
	\centering
	\tiny
	\caption{Key settings of different approaches}
	\label{exp-setting}
	\begin{tabular}{c|c|c|c|c|c}
		\hline\hline
		& \textbf{Cora}                                                                                                   & \textbf{Cite}                                                                                                   & \textbf{Pubmed}                                                                                                 & \textbf{CoauthorCS}                                                                                              & \textbf{OGB-Arxiv}                                                                                                 \\\hline
		\textbf{MoNet}     & \begin{tabular}[c]{@{}c@{}}dropout=0.5\\ lr=0.01\\ hidden=16\end{tabular}                                       & \begin{tabular}[c]{@{}c@{}}dropout=0.5\\ lr=0.01\\ hidden=16\end{tabular}                                       & \begin{tabular}[c]{@{}c@{}}dropout=0.5\\ lr=0.01\\ hidden=16\end{tabular}                                       & \begin{tabular}[c]{@{}c@{}}dropout=0.75\\ lr=0.005\\ hidden=32\end{tabular}                                      & \begin{tabular}[c]{@{}c@{}}dropout=0.75\\ lr=0.005\\ hidden=256\end{tabular}                                       \\\hline
		\textbf{GCN}       & \begin{tabular}[c]{@{}c@{}}dropout=0.5\\ lr=0.01\\ hidden=32\end{tabular}                                       & \begin{tabular}[c]{@{}c@{}}dropout=0.5\\ lr=0.01\\ hidden=32\end{tabular}                                       & \begin{tabular}[c]{@{}c@{}}dropout=0.5\\ lr=0.01\\ hidden=32\end{tabular}                                       & \begin{tabular}[c]{@{}c@{}}dropout=0.5\\ lr=0.005\\ hidden=32\end{tabular}                                       & \begin{tabular}[c]{@{}c@{}}dropout=0.5\\ lr=0.005\\ hidden=256\end{tabular}                                        \\\hline
		\textbf{GraphSAGE} & \begin{tabular}[c]{@{}c@{}}dropout=0.5\\ lr=0.01\\ hidden=16\end{tabular}                                       & \begin{tabular}[c]{@{}c@{}}dropout=0.5\\ lr=0.01\\ hidden=16\end{tabular}                                       & \begin{tabular}[c]{@{}c@{}}dropout=0.5\\ lr=0.01\\ hidden=16\end{tabular}                                       & \begin{tabular}[c]{@{}c@{}}dropout=0.5\\ lr=0.01\\ hidden=32\end{tabular}                                        & \begin{tabular}[c]{@{}c@{}}dropout=0.5\\ lr=0.01\\ hidden=256\end{tabular}                                         \\\hline
		\textbf{JKNet}     & \begin{tabular}[c]{@{}c@{}}dropout=0.5\\ lr=0.005\\ hidden=32\end{tabular}                                      & \begin{tabular}[c]{@{}c@{}}dropout=0.5\\ lr=0.005\\ hidden=32\end{tabular}                                      & \begin{tabular}[c]{@{}c@{}}dropout=0.5\\ lr=0.005\\ hidden=32\end{tabular}                                      & \begin{tabular}[c]{@{}c@{}}dropout=0.5\\ lr=0.005\\ hidden=32\end{tabular}                                       & \begin{tabular}[c]{@{}c@{}}dropout=0.5\\ lr=0.005\\ hidden=256\end{tabular}                                        \\\hline
		\textbf{GAT and variants}       & \begin{tabular}[c]{@{}c@{}}dropout=0.6\\ lr=0.005\\ hidden=8\\ \#hidden heads=8\\ \#output heads=1\end{tabular} & \begin{tabular}[c]{@{}c@{}}dropout=0.6\\ lr=0.005\\ hidden=8\\ \#hidden heads=8\\ \#output heads=1\end{tabular} & \begin{tabular}[c]{@{}c@{}}dropout=0.6\\ lr=0.005\\ hidden=8\\ \#hidden heads=8\\ \#output heads=1\end{tabular} & \begin{tabular}[c]{@{}c@{}}dropout=0.6\\ lr=0.005\\ hidden=32\\ \#hidden heads=8\\ \#output heads=1\end{tabular} & \begin{tabular}[c]{@{}c@{}}dropout=0.75\\ lr=0.002\\ hidden=256\\ \#hidden heads=3\\ \#output heads=3\end{tabular} \\\hline
		\textbf{APPNP}     & \begin{tabular}[c]{@{}c@{}}dropout=0.6\\ lr=0.01\\ hidden=64\end{tabular}                                       & \begin{tabular}[c]{@{}c@{}}dropout=0.6\\ lr=0.01\\ hidden=64\end{tabular}                                       & \begin{tabular}[c]{@{}c@{}}dropout=0.6\\ lr=0.01\\ hidden=64\end{tabular}                                       & \begin{tabular}[c]{@{}c@{}}dropout=0.6\\ lr=0.01\\ hidden=64\end{tabular}                                        & \begin{tabular}[c]{@{}c@{}}dropout=0.75\\ lr=0.01\\ hidden=256\end{tabular}                                        \\\hline
		\textbf{SGC}       & lr=0.05                                                                                                         & lr=0.05                                                                                                         & lr=0.05                                                                                                         & lr=0.05                                                                                                          & lr=0.05                                                                                                            \\\hline
		\textbf{ARMA}      & \begin{tabular}[c]{@{}c@{}}dropout=0.75\\ lr=0.01\\ hidden=64\end{tabular}                                      & \begin{tabular}[c]{@{}c@{}}dropout=0.25\\ lr=0.01\\ hidden=64\end{tabular}                                      & \begin{tabular}[c]{@{}c@{}}dropout=0.25\\ lr=0.01\\ hidden=64\end{tabular}                                      & \begin{tabular}[c]{@{}c@{}}dropout=0.75\\ lr=0.01\\ hidden=64\end{tabular}                                       & \begin{tabular}[c]{@{}c@{}}dropout=0.75\\ lr=0.01\\ hidden=256\end{tabular}                                        \\\hline
		\textbf{GIN}       & \begin{tabular}[c]{@{}c@{}}dropout=0.6\\ lr=0.01\\ hidden=64\\ \end{tabular} & \begin{tabular}[c]{@{}c@{}}dropout=0.6\\ lr=0.01\\ hidden=64\\ \end{tabular} & \begin{tabular}[c]{@{}c@{}}dropout=0.6\\ lr=0.01\\ hidden=64\\ \end{tabular} & \begin{tabular}[c]{@{}c@{}}dropout=0.6\\ lr=0.01\\ hidden=64\\\end{tabular} & \begin{tabular}[c]{@{}c@{}}dropout=0.75\\ lr=0.01\\ hidden=256\\\end{tabular} \\\hline
		\textbf{CAT}       & \begin{tabular}[c]{@{}c@{}}dropout=0.6\\ lr=0.01\\ hidden=8\\ \#hidden heads=8\\ \#output heads=1\end{tabular}  & \begin{tabular}[c]{@{}c@{}}dropout=0.6\\ lr=0.01\\ hidden=8\\ \#hidden heads=8\\ \#output heads=1\end{tabular}  & \begin{tabular}[c]{@{}c@{}}dropout=0.6\\ lr=0.01\\ hidden=8\\ \#hidden heads=8\\ \#output heads=1\end{tabular}  & \begin{tabular}[c]{@{}c@{}}dropout=0.6\\ lr=0.01\\ hidden=8\\ \#hidden heads=8\\ \#output heads=1\end{tabular}   & \begin{tabular}[c]{@{}c@{}}dropout=0.75\\ lr=0.002\\ hidden=256\\ \#hidden heads=3\\ \#output heads=3\end{tabular}\\\hline\hline
	\end{tabular}
\end{table}

\section{Supplementary results}
\subsection{$\mathbf{C}$ pertaining to the relevance of input features}
As mentioned in the manuscript, $\mathbf{C}$ can be any factors representing the node-node relevance that is not considered by the GNN.
To further investigate whether the proposed graph neural network is effective using different interventions external to the GNN, we compute $\mathbf{C}$ based on the cosine similarity (FS) between input features (i.e., $\mathbf X$) and then let different variants of CATs use it to perform semi-supervised node classification and clustering tasks in all the testing datasets.
The performance comparisons between CATs and GATs are summarized in Tables \ref{classification-f} and \ref{clustering-f}.
As the tables show, the performances on both classification and clustering are improved when CATs using similarity in terms of input features (CAT-I-FS and CAT-E-FS) are compared with GAT.
However, they don't perform so well as GAT using concatenation of input feature and structural embeddings in some of the datasets.
As the similarity of input features is similar to the correlation of layer-wise node features internal to the GNN, the performance improvement of CATs is not so evident as that of GAT using concatenation of input feature and structural embeddings.
The obtained results also indicate that considering structural properties in attention-based GNNs may improve their learning performance.

\begin{table*}[htbp]
	\centering
	\caption{Performance comparison on semi-supervised node classification between GATs and CATs using input feature similarity.} 
	\label{classification-f}
	\begin{tabular}{c|ccccc}
		\hline\hline
		&\bf Cora&\bf Cite&\bf Pubmed&\bf CoauthorCS& \bf OGB-Arxiv\\
		\hline
		GAT&83.84 $\pm$ 0.61& 70.36 $\pm$ 0.42& 81.50 $\pm$ 0.47& 92.80 $\pm$ 0.41& 72.39 $\pm$ 0.07\\
		GAT-$k$-Lap&84.10 $\pm$ 0.24&71.18 $\pm$ 0.52 &82.56 $\pm$ 0.30 &92.70 $\pm$ 0.31&72.47 $\pm$ 0.06\\
		GAT-NetMF&84.44 $\pm$ 0.19 &70.94 $\pm$ 0.16 &81.90 $\pm$ 0.33&93.16 $\pm$ 0.27 &72.42 $\pm$ 0.08\\
		GAT-Deep&83.68 $\pm$ 0.67 & 69.70 $\pm$ 0.57 & 80.13 $\pm$ 0.26 & 92.93 $\pm$ 0.17 & 72.79 $\pm$ 0.09 \\
		\hline
		CAT-I-FS& 84.86 $\pm$ 0.30 & 72.54 $\pm$ 0.47 & 82.79 $\pm$ 0.34 & 93.08 $\pm$ 0.24 & 72.57 $\pm$ 0.02\\
		CAT-E-FS & 84.15 $\pm$ 0.23 & 71.68 $\pm$ 0.45 & 82.33 $\pm$ 0.27 & 93.33 $\pm$ 0.23 & 72.52 $\pm$ 0.03\\
		\hline\hline
	\end{tabular}
\end{table*}

\begin{table*}[htbp]
	\centering
	\caption{Performance comparison on semi-supervised node clustering between GATs and CATs using input feature similarity.}
	\label{clustering-f}
	\begin{tabular}{c|ccccc}
		\hline\hline
		&\bf Cora&\bf Cite&\bf Pubmed&\bf CoauthorCS& \bf OGB-Arxiv\\
		\hline
		GAT&81.39 $\pm$ 0.18& 69.20 $\pm$ 0.28& 80.88 $\pm$ 0.33& 90.09 $\pm$ 0.15& 76.04 $\pm$ 0.38\\
		GAT-$k$-Lap&80.66 $\pm$ 0.31 & 69.56 $\pm$ 0.34 & 81.59 $\pm$ 0.09 & 89.83 $\pm$ 0.18 & 76.21 $\pm$ 0.06 \\
		GAT-NetMF &81.75 $\pm$ 0.26 & 68.96 $\pm$ 0.21 & 81.74 $\pm$ 0.18 & 89.85 $\pm$ 0.21 & 76.06 $\pm$ 0.07\\
		GAT-Deep &81.08 $\pm$ 0.41 & 68.27 $\pm$ 0.06 & 80.55 $\pm$ 0.11 & 89.70 $\pm$ 0.27 & 76.91 $\pm$ 0.15\\
		\hline
		CAT-I-FS & 81.83 $\pm$ 0.23 & 70.36 $\pm$ 0.43 & 81.89 $\pm$ 0.37 & 90.20 $\pm$ 0.17 & 76.33 $\pm$ 0.03\\
		CAT-E-FS & 81.50 $\pm$ 0.26 & 70.37 $\pm$ 0.43 & 81.91 $\pm$ 0.34 & 90.01 $\pm$ 0.19 & 76.18 $\pm$ 0.33\\ 
		\hline\hline
	\end{tabular}
\end{table*}

\begin{table}[]
	\small
	\centering
	\caption{Model comparison between GAT and CAT using different learning strategies}
	\label{complexity}
	\begin{tabular}{c|c|ccccc}
		\hline\hline
		&                   & \textbf{Cora} & \textbf{Cite} & \textbf{Pubmed} & \textbf{CoauthorCS} & \textbf{OGB-Arxiv} \\\hline
		\multirow{2}{*}{\textbf{GAT}}      & \# Parameters     & 92430         & 237644        & 32454           & 1746974             & 384280             \\
		& Space consumption & 1.1GB         & 1.2GB         & 1.2GB           & 2.3GB               & 5.6GB              \\\hline
		\multirow{2}{*}{\textbf{CAT-I-SC}} & \# Parameters     & 111285        & 257505        & 91504           & 2021484             & 7158006            \\
		& Space consumption & 0.8GB         & 1.0GB         & 1.3GB           & 2.5GB               & 21.4GB             \\\hline
		\multirow{2}{*}{\textbf{CAT-I-MF}} & \# Parameters     & 111285        & 257505        & 91504           & 2021484             & 7158006            \\
		& Space consumption & 1.0GB         & 1.1GB         & 1.3GB           & 2.5GB               & 6.3GB              \\\hline
		\multirow{2}{*}{\textbf{CAT-E-SC}} & \# Parameters     & 111267        & 257487        & 91486           & 2021466             & 7158002            \\
		& Space consumption & 1.1GB         & 1.3GB         & 1.4GB           & 2.5GB               & 21.4GB             \\\hline
		\multirow{2}{*}{\textbf{CAT-E-MF}} & \# Parameters     & 111267        & 257487        & 91486           & 2021466             & 7158002            \\
		& Space consumption & 1.2GB         & 1.3GB         & 1.4GB           & 2.5GB               & 6.6GB\\\hline\hline                          
	\end{tabular}
\end{table}

\subsection{Experiment on space complexity}
When using the proposed CATs to learn representations from graph data, we allow the learning of $\mathbf C_{ij}$ (i.e., Eq. (2) or (3) in the manuscript) to be simultaneously done with the training of the neural network.
To do so, the loss function of CATs can be simply reformulated as $L = L_{predict} + \lambda L_{ext}$, where $L_{predict}$ is the task-specific loss, $L_{ext}$ is the loss for learning all $\mathbf C_{ij}$, and $\lambda$ is the balancing parameter.
To further reduce the computational burden in the training stage, we use a low-dimensional matrix $\mathbf V$ (with dimension $N$-by-$C$ in our experiments) to approximate $\mathbf C$ (see Eq. (2) or (3) in the manuscript).
Aiming at investigating whether the proposed CATs utilizing the aforementioned training strategy can be used in massive graph datasets, we record the number of model parameters of CATs and their memory consumption in all the testing datasets and compare it with the closely related attention-based GNN, GAT.
The corresponding results are summarized in Table \ref{complexity}.
Compared with GAT, CATs use more learnable parameters to complete the task of representation learning in different datasets.
Such growth of parameters in CATs is mainly because CATs have to simultaneously learn structural interventions (e.g., $\mathbf {C}_{ij}$ in Eq. (2) or (3) in the manuscript) for the computation of conjoint attentions.
As for the space complexity, how much video memory is used by CATs jointly determined by the graph neural architecture and the learning of $\mathbf {C}_{ij}$.
When a simple but effective paradigm, e.g., MF shown in Eq. (2) in the manuscript is used for learning $\mathbf C_{ij}$, the memory consumption of CATs is very close to that of GAT, although they have more parameters involved into the back-propagation process.
For example, in OGB-Arxiv, which is the largest dataset used in our experiments, CATs take only 1GB more than GAT does to learn representations.
Given the competitive results on space complexity, it is said that the proposed Conjoint Attentions and corresponding CATs can be used in massive graph data for effective representation learning.

\subsection{Sensitivity test on $\lambda$}
\begin{figure}[!htb]
	\centering
	\begin{subfigure}[b]{0.24\linewidth}
		\includegraphics[width=\textwidth]{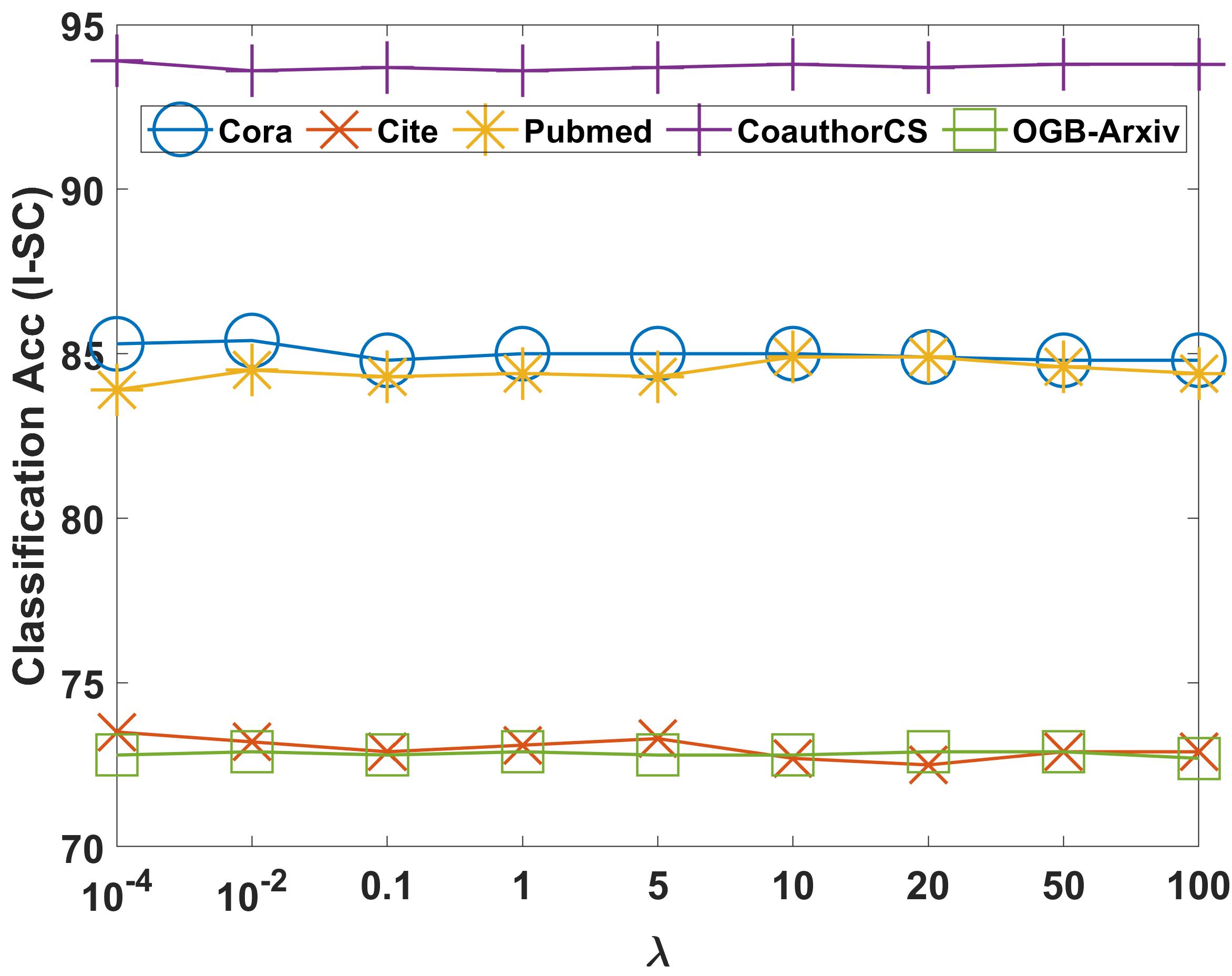}
	\end{subfigure}
	\begin{subfigure}[b]{0.24\linewidth}
		\includegraphics[width=\textwidth]{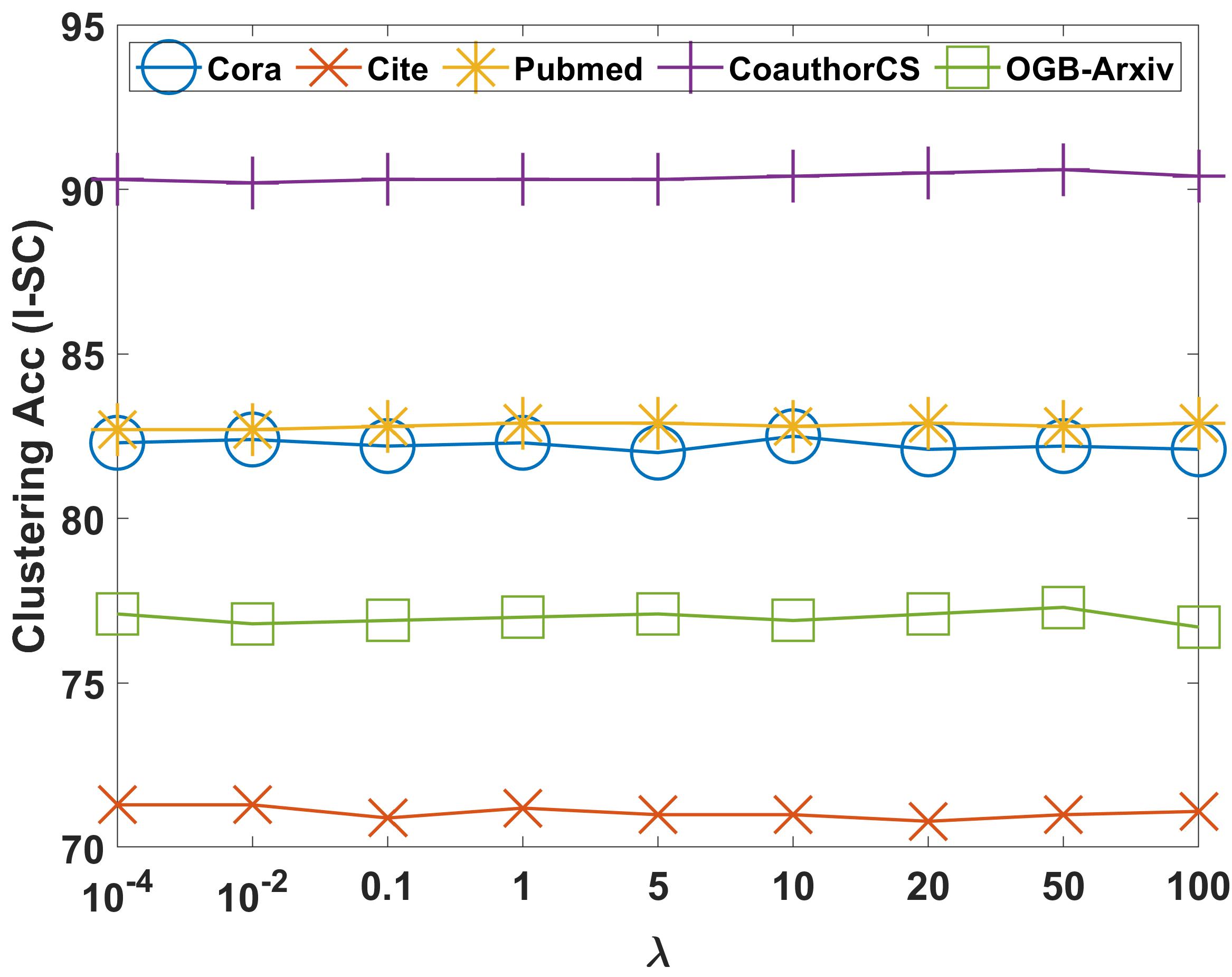}
	\end{subfigure}
	\begin{subfigure}[b]{0.24\linewidth}
		\includegraphics[width=\textwidth]{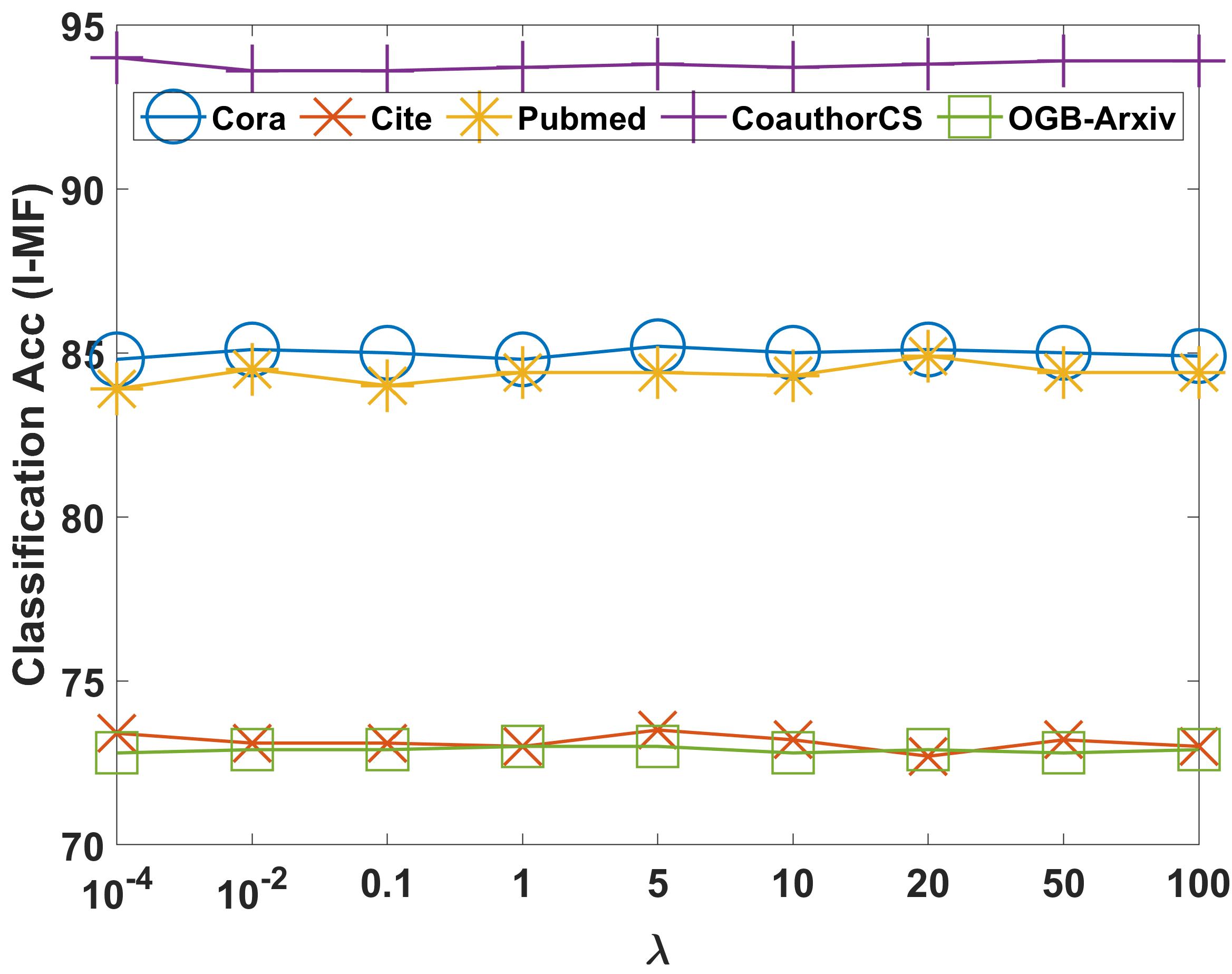}
	\end{subfigure}
	\begin{subfigure}[b]{0.24\linewidth}
		\includegraphics[width=\textwidth]{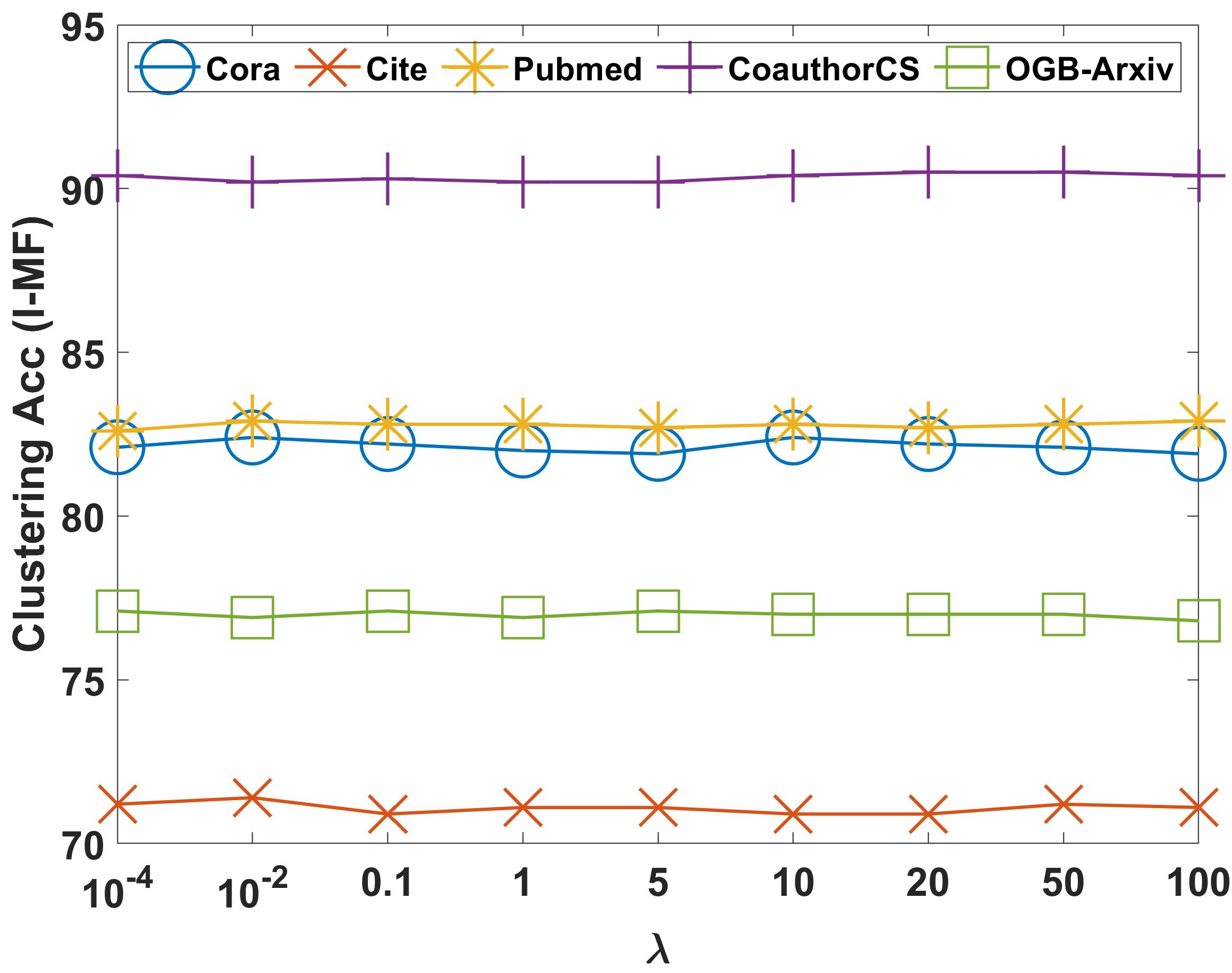}
	\end{subfigure}
	\begin{subfigure}[b]{0.24\linewidth}
		\includegraphics[width=\textwidth]{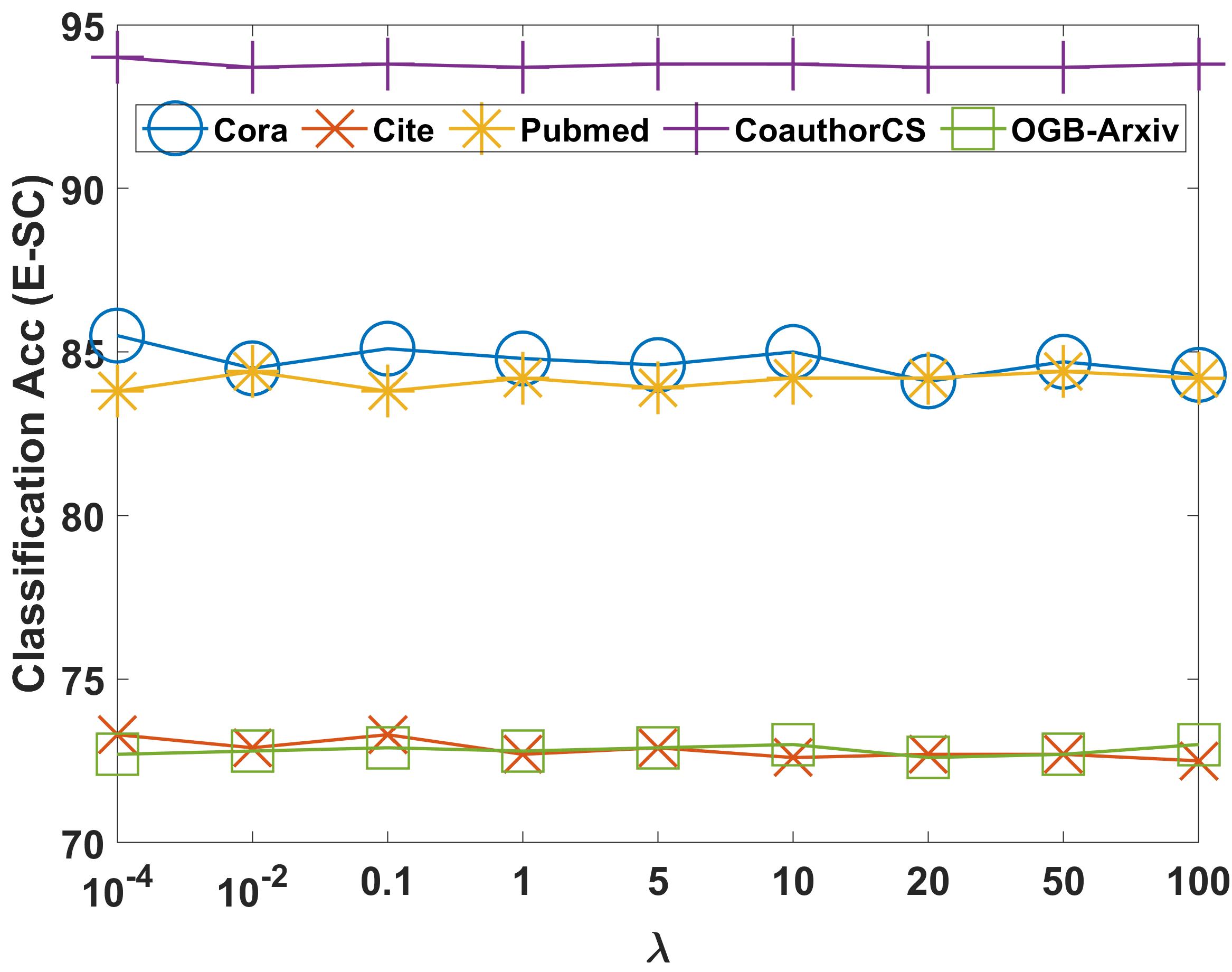}
	\end{subfigure}
	\begin{subfigure}[b]{0.24\linewidth}
		\includegraphics[width=\textwidth]{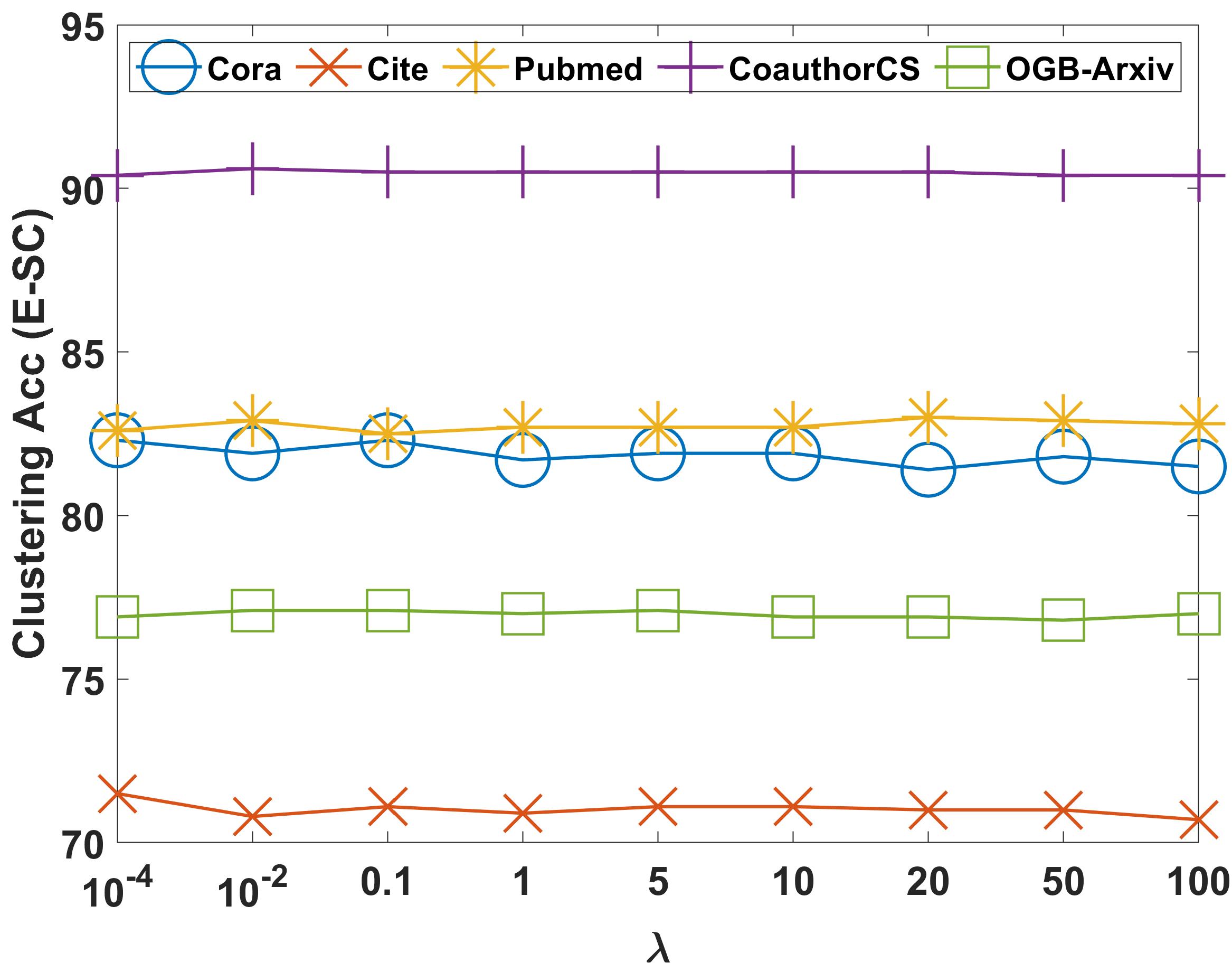}
	\end{subfigure}
	\begin{subfigure}[b]{0.24\linewidth}
		\includegraphics[width=\textwidth]{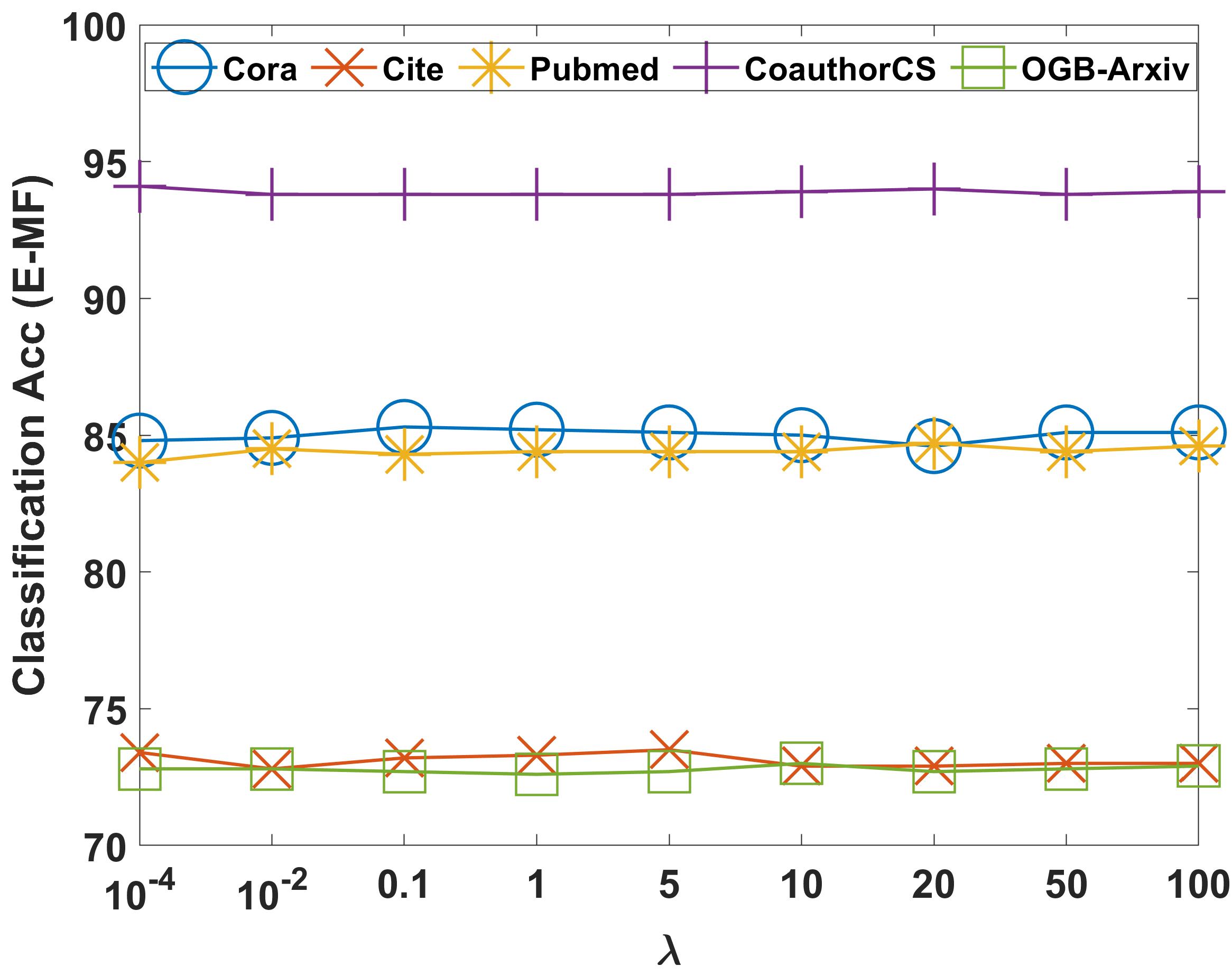}
	\end{subfigure}
	\begin{subfigure}[b]{0.24\linewidth}
		\includegraphics[width=\textwidth]{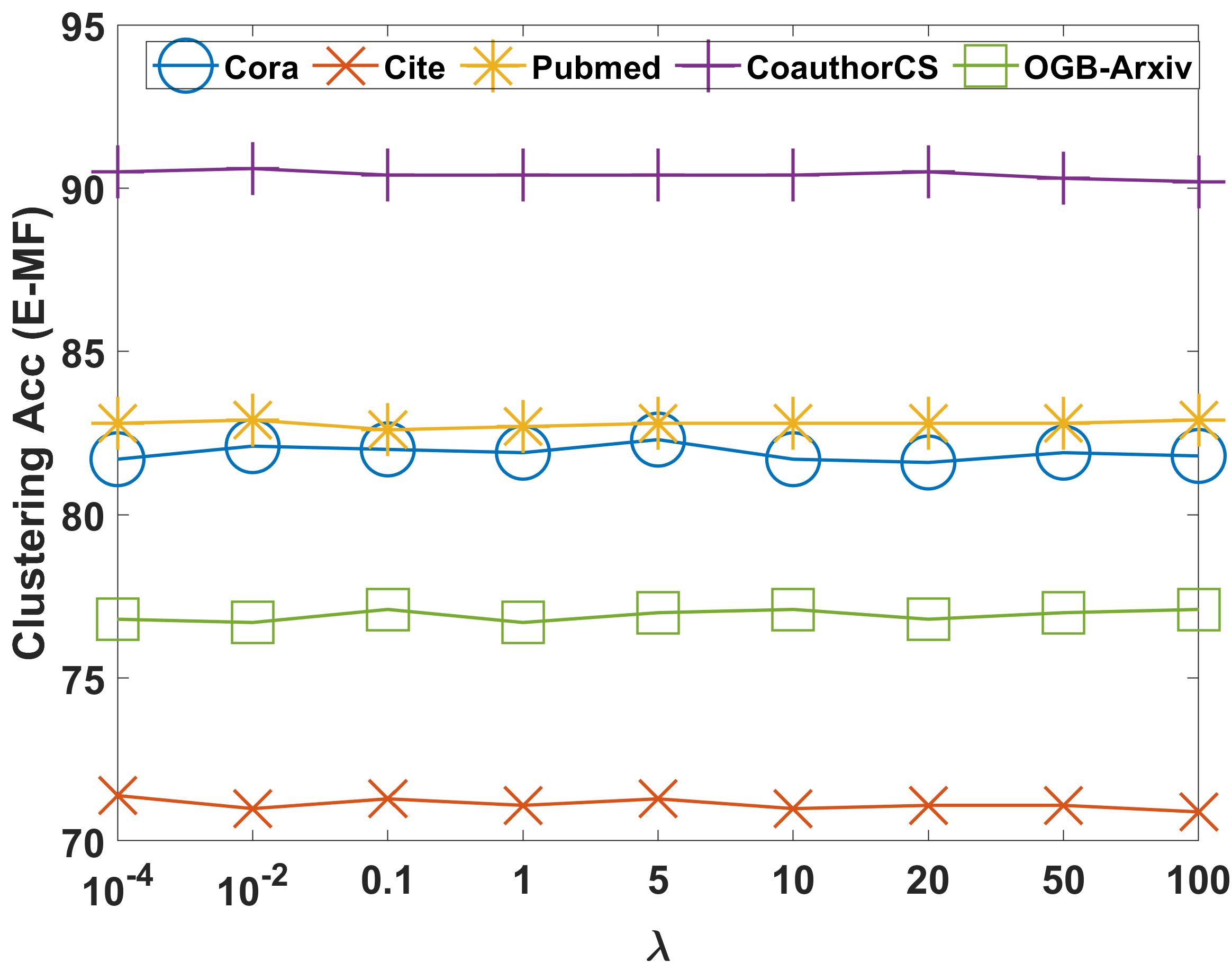}
	\end{subfigure}
	\caption{Sensitivity test on $\lambda$}\label{sens}
\end{figure}

As structural interventions ($\mathbf C$) are learnable, we allow the learning of $\mathbf C$ to be simultaneously done with the training of CATs.
Thus, the loss function of CATs becomes $L = L_{predict} + \lambda L_{ext}$, where $L_{predict}$ is the task-specific loss (e.g., cross entropy loss for node classification), $L_{ext}$ can be the sum of all possible items shown in either Eq. (2) or (3), and $\lambda$ is a balancing parameter for controlling the relative significance of learning $\mathbf C_{ij}$.
To investigate the model sensitivity against $\lambda$, we set $\lambda = [10^{-4}, 10^{-2}, 10^{-1}, 1, 5, 10, 20, 50, 100]$ and run CATs in all datasets to test their performance on different settings of $\lambda$.
The results have been plotted in Fig. \ref{sens}. 
As depicted, all the versions of CATs may perform robustly under a wide range settings of $\lambda$.

Obtaining the presented results might be due to the following reasons.
In this paper, we mostly use semi-supervised learning tasks to test the effectiveness of different approaches.
This means only a very low proportion of node labels (e.g., 140 out of 2708 in dataset Cora) are used to compute $L_{predict}$ and consequently only the gradients related to this small number of labeled nodes will be computed in the back propagation stage of CATs in each training epoch.
Besides, a proportion of the gradients related to the labeled nodes might still be set to zero due to the dropout mechanism in each layer of the GNN.
Given the mentioned property of learning tasks and the dropout mechanism used by GNN, in each training epoch, the gradients related to $L_{predict}$ contribute very limitedly to the learning of $\mathbf C_{ij}$.
In contrast, $L_{ext}$ sums all the $\mathbf C_{ij}$s and we do not apply dropout on the training of $L_{ext}$, i.e., the learning of $\mathbf C_{ij}$.
As a result, the learning of $\mathbf C_{ij}$ is dominantly determined by $L_{ext}$, i.e., Eqs. (2) or (3) in the manuscript, and the proposed CATs are less sensitive to the changes of $\lambda$. Changing the settings of $\lambda$ may only affect the convergence of learning $\mathbf C_{ij}$.
For simplicity, we set $\lambda = 0.01$ for CATs in our experiments.

\section{Computational complexity}
The computational complexity of the proposed CATs is mainly determined by the computation of attention layers and structural interventions.
For the computation of each attention head in the graph neural network, its complexity for learning $D^{l+1}$ features for each node is same to the classical GAT, and can be represented as $O(ND^lD^{l+1}+(\vert E\vert + e) D^{l+1})$, where $e$ represents average degree of the nodes. If there are $K$ attention heads used, the complexity becomes $O(K(ND^lD^{l+1}+(\vert E\vert + e) D^{l+1}))$.
As for the structural interventions for computing Conjoint Attentions, i.e., $\mathbf C_{ij}$, the complexity depends on its learning method.
For example, when either matrix factorization or subspace learning method (Eq. (2) or (3) in the manuscript) is used, the complexity for learning each $\mathbf C_{ij}$ in each epoch is approximately $O(2(C+1)N+C)$, where we assume the dimension of $\mathbf V$ that is used for approximating $\mathbf C$ is $N$-by-$C$.

\end{document}